\documentclass[twoside,11pt]{article} 

\usepackage{jmlr2e}

\PassOptionsToPackage{numbers, compress}{natbib}
\usepackage{natbib}
\usepackage{times}
\usepackage{epsfig}
\usepackage{graphicx}
\usepackage{amsmath}
\usepackage{amssymb}
\usepackage{xcolor}
\usepackage{booktabs}
\usepackage{subfigure}
\usepackage{wrapfig}
\usepackage[accsupp]{axessibility}  

\definecolor{jamescolor}{rgb}{0.858, 0.188, 0.478}
\definecolor{chocolor}{cmyk}{0, 0.7808, 0.4429, 0.1412}

\newif\ifcomments
\commentstrue

\usepackage{caption}
\ShortHeadings{Causal Scene BERT: Improving object detection by searching for challenging groups of data}{Resnick and Litany and Kar and Kreis and Lucas and Cho and Fidler}
\firstpageno{1}

\begin{document}

\title{Causal Scene BERT: Improving object detection by searching for challenging groups of data}

\author{
  \name Cinjon Resnick \email cinjon@nyu.edu \\
  \addr New York University - CILVR, 60 Fifth Ave, New York, NY, 10011 \\
  \AND
  \name Or Litany \email olitany@nvidia.com \\
  \addr NVIDIA - 2788 San Tomas Expy, Santa Clara, CA 95051 \\
  \AND
  \name Amlan Kar \email amlan@cs.toronto.edu \\
  \addr NVIDIA; University of Toronto; Vector Institute \\
  \AND
  \name Karsten Kreis \email kkreis@nvidia.com \\
  \addr NVIDIA \\
  \AND
  \name James Lucas \email jalucas@nvidia.com \\
  \addr NVIDIA \\
  \AND 
  \name Kyunghyun Cho \email kyunghyun.cho@nyu.edu \\
  \addr New York University - CILVR, 60 Fifth Ave, New York, NY, 10011 \\
  \AND 
  \name Sanja Fidler \email sfidler@nvidia.com \\
  \addr NVIDIA; University of Toronto; Vector Institute
  }

\editor{Kevin Murphy and Bernhard Sch{\"o}lkopf}

\maketitle

\begin{abstract}

Modern computer vision applications rely on learning-based perception modules parameterized with neural networks for tasks like object detection. These modules frequently have low expected error overall but high error on atypical groups of data due to biases inherent in the training process. In building autonomous vehicles (AV), this problem is an especially important challenge because their perception modules are crucial to the overall system performance. After identifying failures in AV, a human team will comb through the associated data to group perception failures that share common causes. 
More data from these groups is then collected and annotated before retraining the model to fix the issue. In other words, error groups are found and addressed in \emph{hindsight}. Our main contribution is a pseudo-automatic method to discover such groups in \emph{foresight} by performing causal interventions on simulated scenes. To keep our interventions on the data manifold, we utilize masked language models.
We verify that the prioritized groups found via intervention are challenging for the object detector and show that retraining with data collected from these groups helps inordinately compared to adding more IID data. We also plan to release software to run interventions in simulated scenes, which we hope will benefit the causality community. 

\end{abstract}




\maketitle


\section{Introduction}
\label{sec:intro}

To deploy robotic systems such as autonomous road vehicles, it is vital that they are robust and safe. An important aspect of safety is handling unusual scenarios. Current data-driven approaches trained to minimize expected error are sensitive to imbalanced data distributions. As a result, models with low expected error can still exhibit large errors on atypical groups of data that are nonetheless important for safe driving.
The status quo approach to finding these groups in the AV 
stack operates in \emph{hindsight} by analyzing real-world scenes requiring driver intervention or by feeding replayed or simulated scenes to a model and finding those that result in poor performance. Advanced techniques may use adversarial attacks to actively find failures~\citep{xie2017adversarial,DBLP:journals/corr/AthalyeS17,wang2021advsim, rempe2021generating}. In all cases, the found data is fed back into the retraining process. While this improves the model, a notable problem remains --- without knowing the underlying cause of a failure, it is impossible to ensure that the problem is adequately resolved. 
To identify the causal factors in the failures, human experts typically comb through the data and group commonalities, an expensive and time-consuming procedure.



\begin{figure*}[t!]
    \centering
    \includegraphics[width=0.9\textwidth, trim=50 35 40 35, clip]{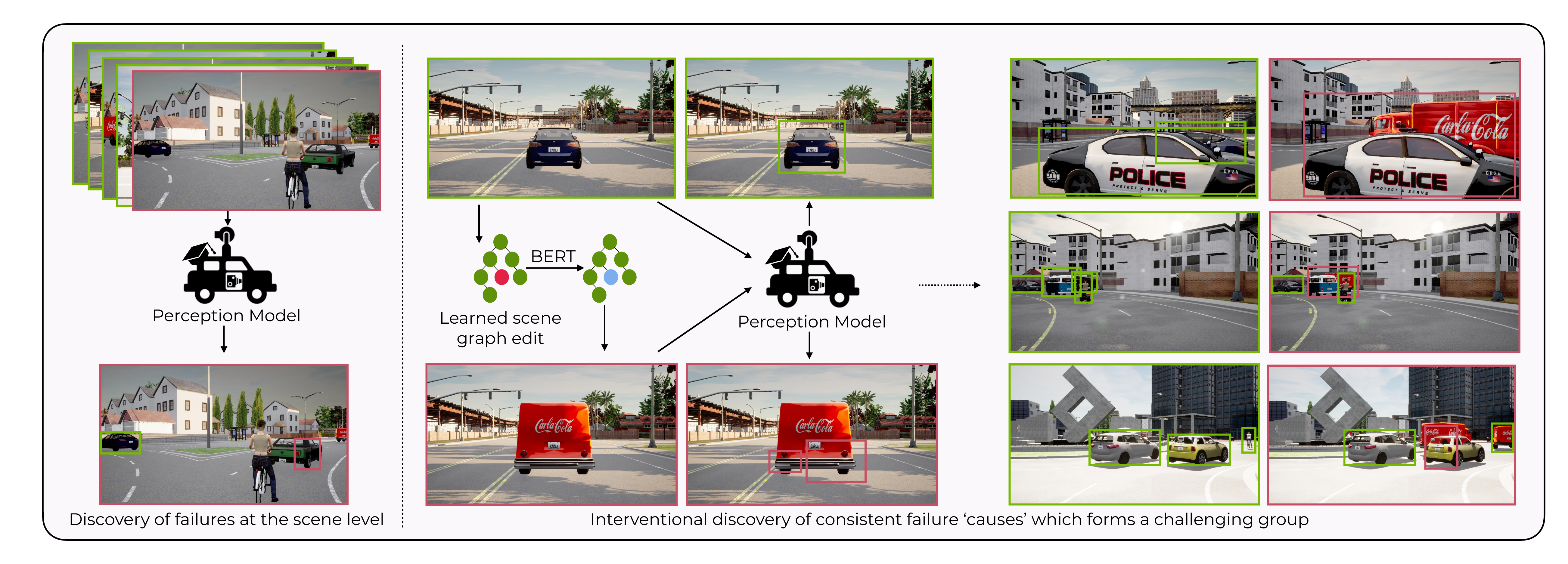}
    \caption{Instead of retrospectively discovering individual failure cases for perception, we actively search for causal interventions (edits) to existing scenes that consistently result in perception failures. The middle shows an example of a single intervention causing perception failure, which we attribute to the intervention, as opposed to the left where a combinatorial set of factors could explain the error. Consistent failures through this type of intervention constitute a challenging group for the perception model as seen on the right.}
    \label{fig:teaser}
    \vspace{-6mm}
\end{figure*}

We propose an alternative method to discover potential 
failures in \emph{foresight} as shown in Figure~\ref{fig:teaser}. Instead of finding failures from previously collected data, we perform interventions on existing data to find those interventions that are detrimental to the performance of an AV 
stack. We focus on perception, and object detection specifically, in this work. We identify interventions that consistently cause performance drops as challenging groups. 
Concretely, consider a scene where a truck was not detected. Many explanations exist, ranging from the scene composition to the weather conditions to the way the light reflects off of a puddle and into the camera. The actual cause is unclear. If we however arrived at this scene counterfactually, by performing a single intervention on another scene, e.g. changing a car to the truck, we now have some clue that the underlying causal error is related to the truck itself. We can duplicate this intervention across many scenes and see if it consistently remains a problem. While the \textit{exact} cause is still opaque, the proposed method provides automatic insight into what interventions cause consistent errors without collecting new data to analyze or manual scrubbing of failures.

Performing such interventions requires the ability to manipulate scenes and re-render images. We demonstrate this in simulation, although recent advances~\citep{ost2020neuralscenegraphs} show promise in migrating our approach to real-world scenes. We assume access to a scene graph representation of the underlying scene on which we perform interventions. These interventions include changing agent properties like position, rotation, or asset type, as well as global weather conditions. While many interventions can potentially fail the detector, not all are useful. A scene with a flying truck could drop perception performance, but it is unlikely to occur in the real world. Ideally, interventions should be from the data distribution. We achieve this by training a density model of scenes (represented as flattened scene graphs) using a masked language model (MLM), a keystone in modern natural language processing pipelines. Taking interventions using the MLM amounts to masking a part of the scene graph and re-sampling from the predicted distribution.



Our work focuses on 2D object detection from input images of driving scenes. We verify that the prioritized groups we find via intervention are indeed challenging for the base object detector and show that retraining with data collected from these groups helps inordinately compared to adding more IID data. We additionally confirm our hypothesis that interventions on the data distribution are preferred vis a vis data efficiency by comparing against random interventions. The latter are confounded by their propensity to stray from the data distribution. We compare these results against an important baseline we call `Cause-agnostic Data Collection', which are scenes for which the model performs poorly according to the same custom scoring function used for the interventions. Finally, we examine what happens when we take a second intervention using the MLM and find new veins in which we could mine specific problematic groups, suggesting that there is room to continue this process.

\textbf{Our primary contribution is a novel method using Masked Language Models (MLMs) to intervene on scene graphs of simulated scenes to causally uncover semantic groups of data upon which a detection model is likely to fail}. Unlike sporadic failures, our found failure groups provide insight into the model's weaknesses and help us systematically improve the model.

\section{Background}
\label{sec:background}

\paragraph{Notation}

Our objective is to ascertain the capabilities of a given object detection model $\phi$. We represent a scene $x$ as a triplet $x = (G, I, L)$ of a scene graph (includes the camera parameters), scene image, and a set of bounding box labels, respectively. We flatten and discretize $G$ to get the corresponding sequence $S \in \mathbb{N}^{O(d)}$ where $d$ is the variable number of objects in the scene. The scene image $I \in \mathbb{R}^3$ is the RGB image of the scene as observed by the ego car and is deterministically defined by $G$. The label $L$ is a set of ground truth bounding boxes $l_k \in \mathbb{R}^4$, where $k < d$ is the number of objects to identify in the scene. 
Scenes are drawn from a distribution $p_{R}(x)$ dictated by the scene generation process $R$. Examples of $R$ include driving scenes from a particular city or simulations from AV simulators. We also define a per-example scoring function $f: (\phi, I, L) \rightarrow y \in [0, 1]$ as well as a threshold $\tau$ with which to gauge whether an intervention was detrimental.

\paragraph{Scene Graphs} are 3D world representations, with nodes corresponding to entities and edges to hierarchical relationships between nodes, where the hierarchy is determined by physical presence (e.g. road is a parent of the vehicles on the road). Entities include the vehicles and pedestrians, the weather, the ego agent, and the camera parameters. Each node has associated attributes, exemplified by continuous rotation, continuous position, discrete asset type, etc. 

\paragraph{Object Detection} in images reached a milestone with Faster RCNN~\citep{ren2016faster}. We use their approach as the representative state of the art detection module via the Detectron2 library~\citep{wu2019detectron2}. 

\paragraph{Simulation} is crucial to our method. We need a simulator that can initialize from $G$, have satisfactory traffic policies for autonomous vehicles, and return the current $G$ on command. The chosen CARLA~\citep{dosovitskiy_2019} simulator satisfies these constraints and is ubiquitous in the field.

\paragraph{Masked Language Models (MLM)} are density models for sequential data. \citet{devlin2019bert} showed their tremendous efficacy in language generation. They are trained by receiving sequences of discrete tokens, a few of which are masked, and predicting what tokens are in the masked positions. Through this process, they learn the data distribution~\citep{DBLP:journals/corr/abs-2009-10195,DBLP:journals/corr/abs-1902-04094} of those sequences. At inference, they are fed a sequence with a chosen token masked and replace the mask with their prediction. We perform causal intervention on scenes by asking a pre-trained MLM to re-sample a single position from a scene graph - see Section~\ref{sec:experiments-setup} for details.

\section{Method}
\label{sec:method}

We aim to improve object detection models by utilizing the advantages of AV simulators over collecting real world data, namely that they quickly synthesize scenes in parallel; that we have fine control over the synthesis; and that they grant us supervisory labels automatically. A naive approach is to generate lots of random scenes, test detection on those scenes, and set aside the hard ones for retraining. A more advanced one is to use adversarial techniques to find hard scenes. Both approaches share two downsides: a) we are much more interested in scenes drawn from a distribution that is similar to the distribution of real-world vehicle scenes and b) there is a combinatorial challenge of understanding what in the scenes was the problem; only if we know why the error is happening can we find test scenes having similar challenges and thus understand if the issue is fixed after retraining.

\begin{figure}[!t]
    \centering
    \includegraphics[width=0.9\textwidth]{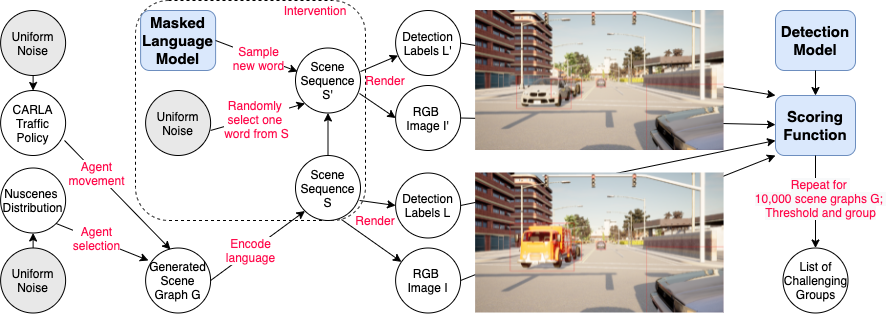}
    \caption{\textbf{A complete diagram of our approach}: We intervene on the scene by performing transformations like the pictured yellow truck becoming a white car and then evaluating the delta change in the object detection model's efficacy. The interventions are guided by a trained MLM. Repeat $N$ times and group the scores to attain an ordered list of challenging groups drawn from vehicle type, weather, and rotation.
    }
    \label{fig:scenegraphconversion}
\end{figure}

\begin{figure}[!h]
    \centering
    \begin{minipage}{0.45\textwidth}
        \centering
        \captionsetup{labelformat=empty}        
        \caption{Ground truth boxes}
        \includegraphics[width=\textwidth]{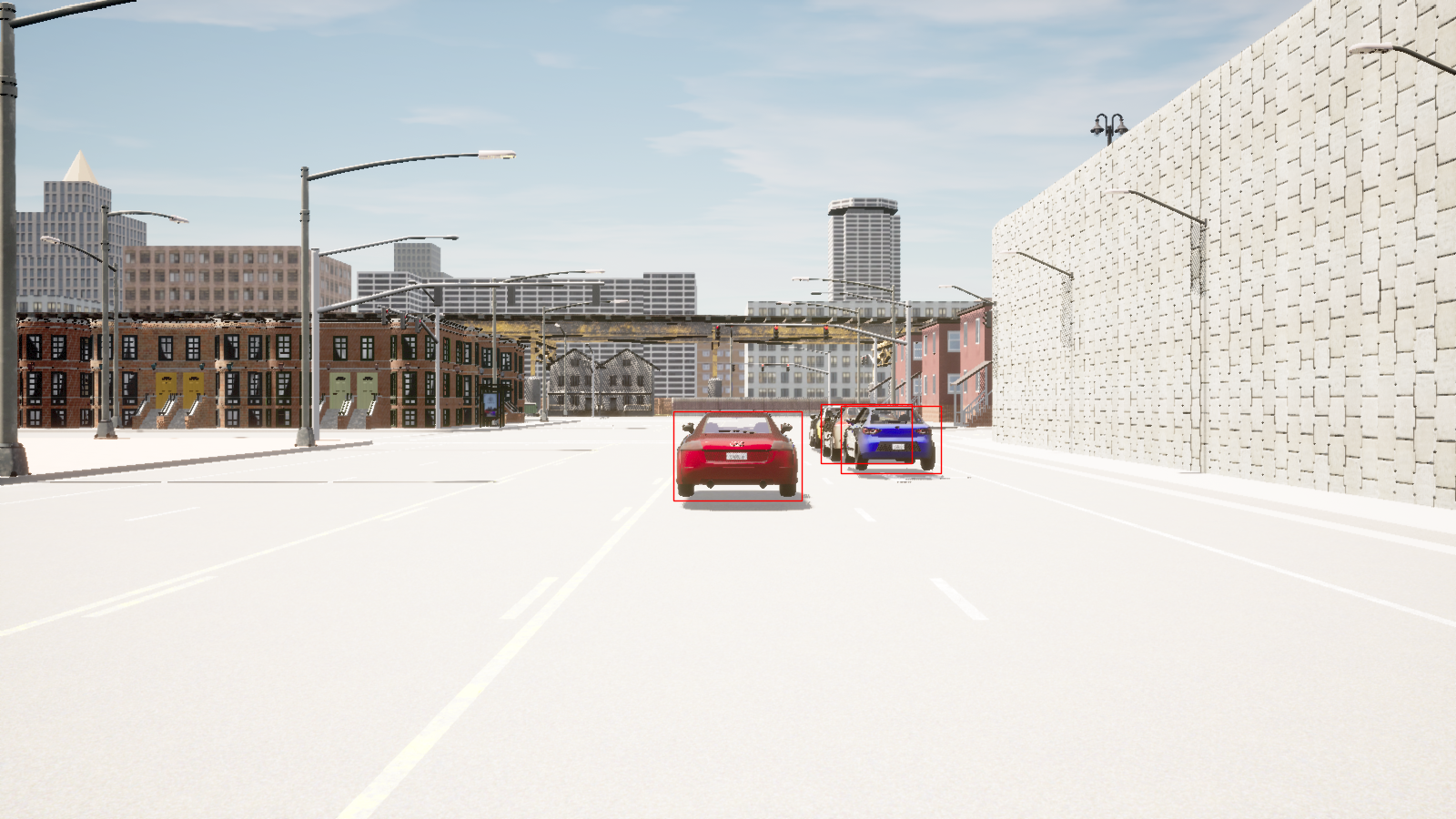} 
    \end{minipage}\hfill
    \begin{minipage}{0.45\textwidth}
        \centering
        \captionsetup{labelformat=empty}
        \caption{Object detector's output}
        \includegraphics[width=\textwidth]{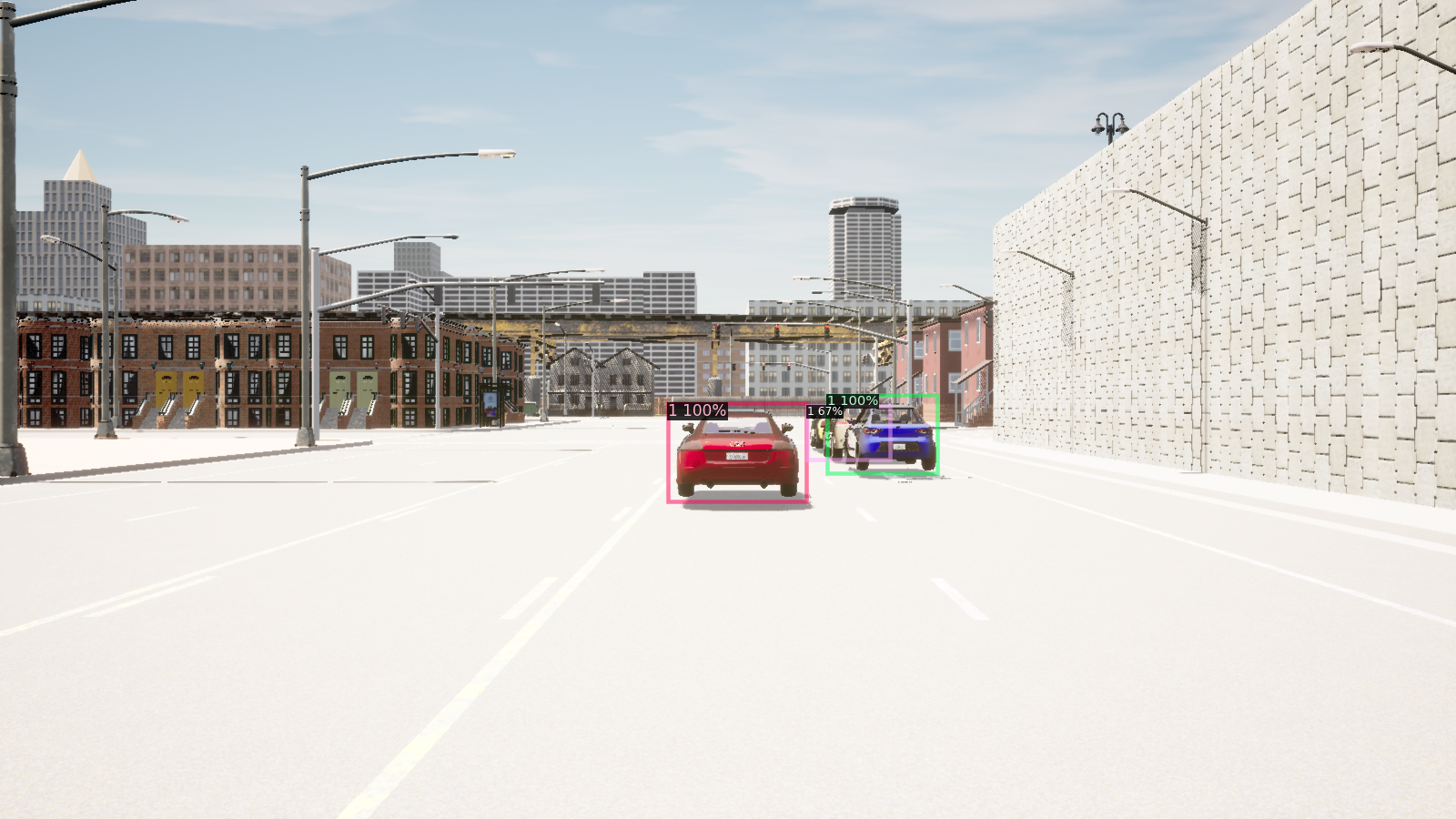} 
    \end{minipage}
    
    \begin{minipage}{0.45\textwidth}
        \centering
        \includegraphics[width=\textwidth]{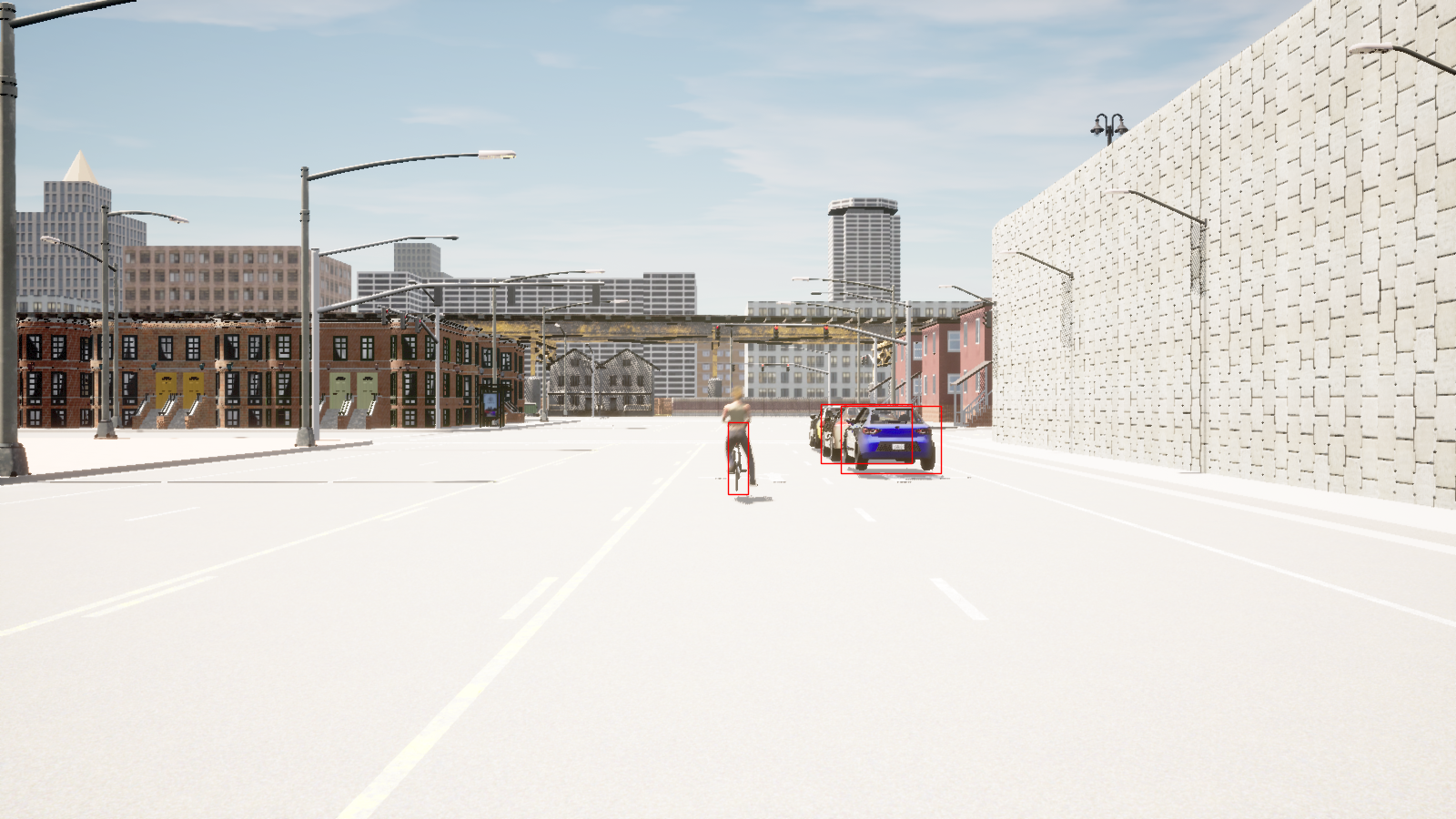} 
    \end{minipage}\hfill
    \begin{minipage}{0.45\textwidth}
        \centering
        \includegraphics[width=\textwidth]{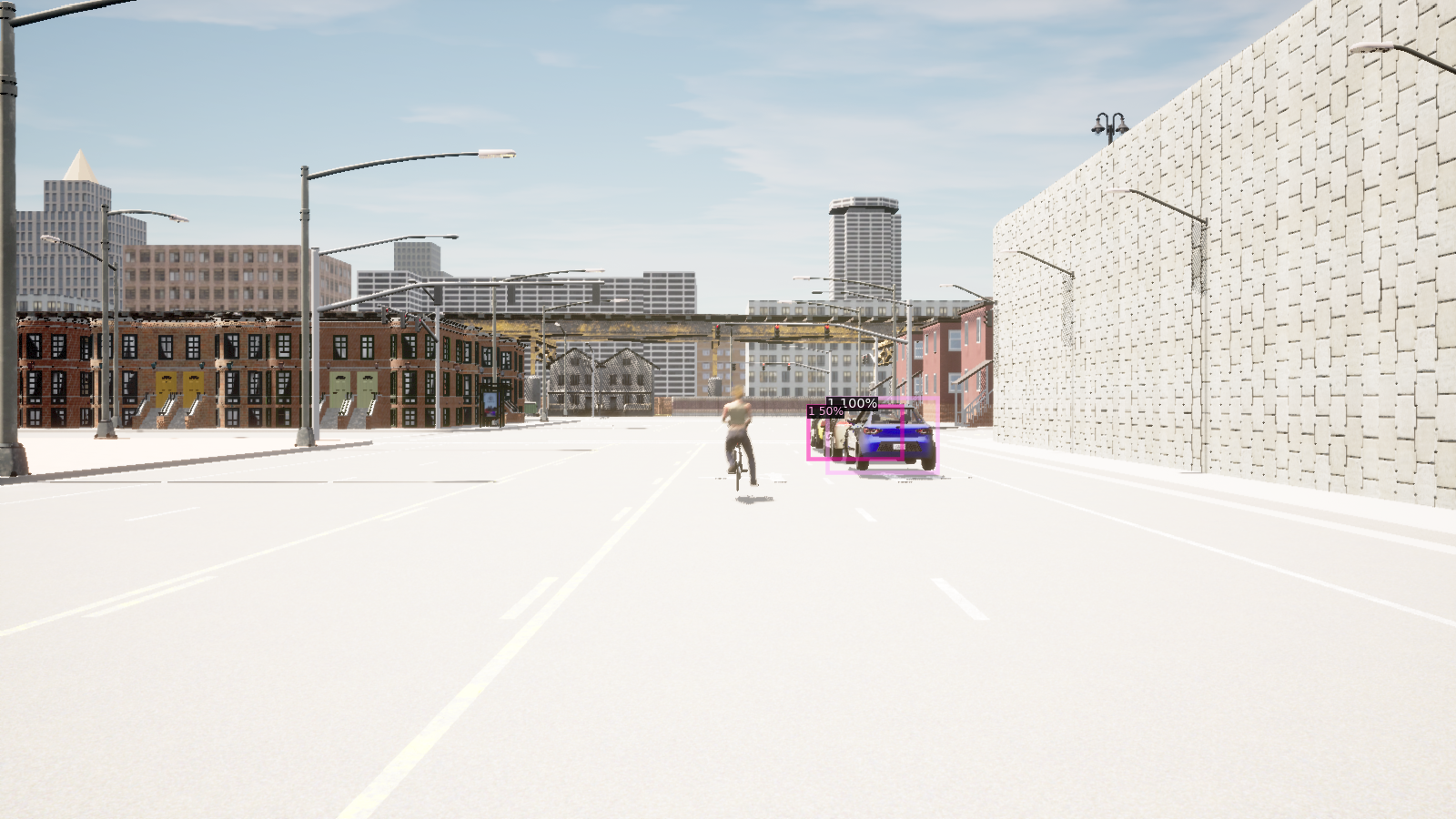} 
    \end{minipage}   
    
    \begin{minipage}{0.45\textwidth}
        \centering
        \includegraphics[width=\textwidth]{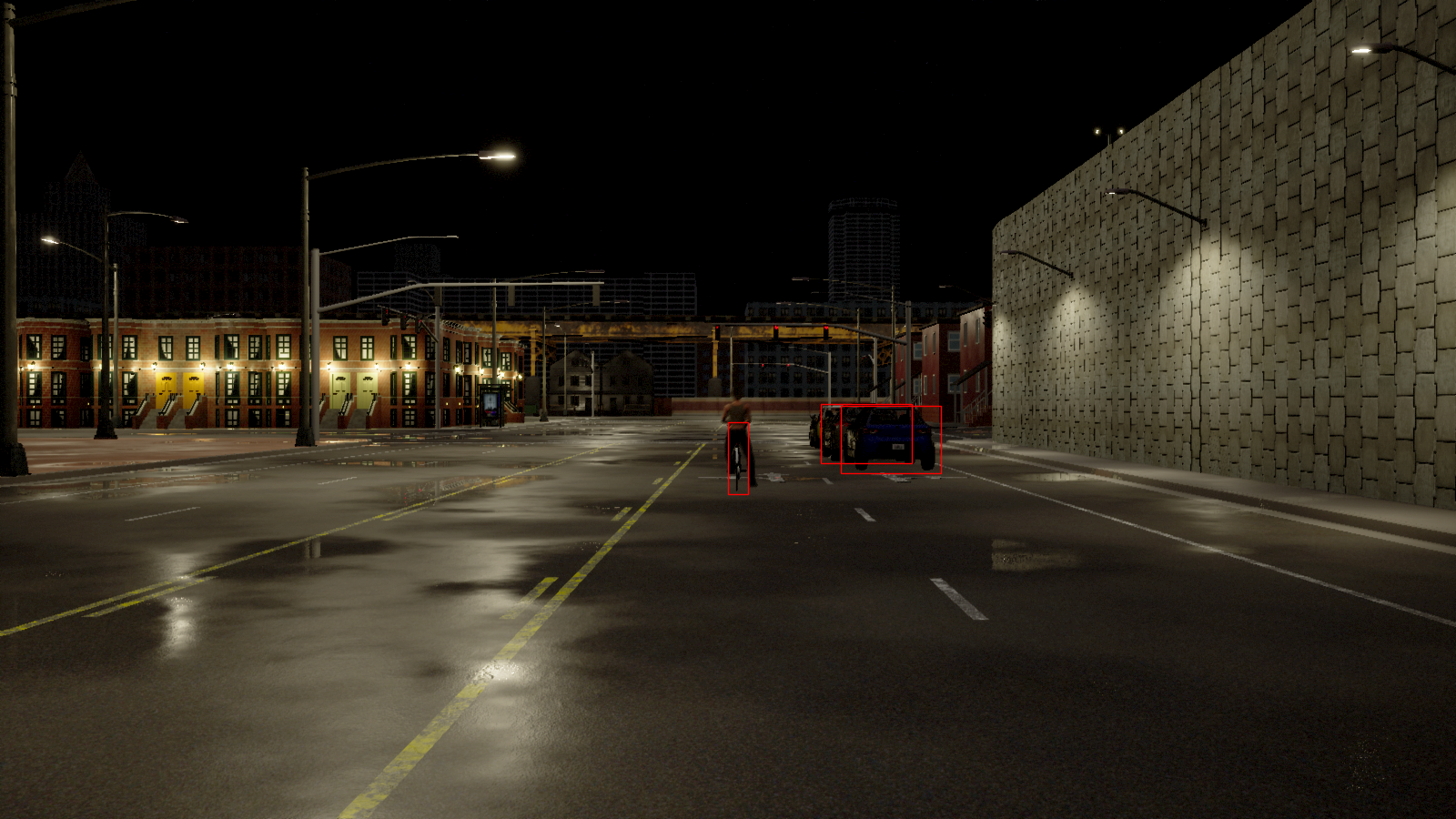} 
    \end{minipage}\hfill
    \begin{minipage}{0.45\textwidth}
        \centering
        \includegraphics[width=\textwidth]{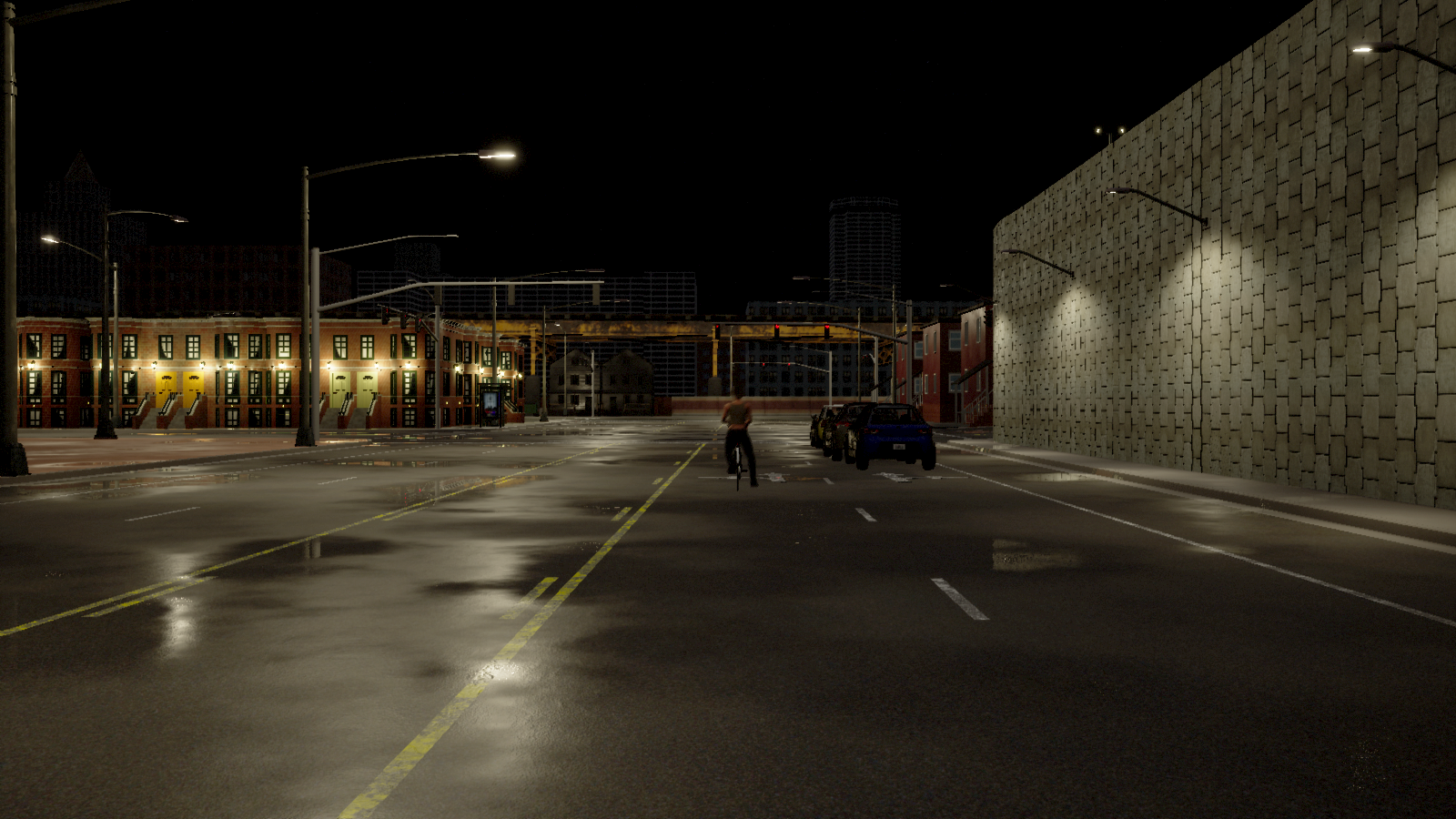} 
    \end{minipage} 
    \captionsetup[figure]{labelformat=default}
    \caption{\textbf{Interventions taken by the MLM}. The first row is the original scene, the second after an intervention changing the red car to a biker, and the third after an intervention on the weather. The left side shows ground truth and the right shows the detector's predictions. Observe that the model's predictions deteriorate from being essentially perfect to missing the biker to missing every object. See Figure~\ref{fig:intervention-example-appendix} in the Appendix for a rotation example.}
    \label{fig:intervention-example}
\end{figure}

We propose an efficient procedure that tackles both concerns. We find hard groups of data for a trained model $\phi$ by taking interventions on scene graphs with an MLM pre-trained on natural scene distributions. The resulting scenes are grouped according to their generating intervention type. We assess the model performance on each group with our surrogate scoring function $f$. The rationale behind this procedure is that solely identifying challenging \textit{scenes} does not provide insight into how to improve $\phi$. However, asserting that a type of \textit{intervention} is consistently challenging narrows greatly where the model's difficulties lay. After finding challenging groups, we utilize hard negative mining~\citep{DBLP:journals/corr/abs-1709-02940,DBLP:journals/corr/KumarHC0D17,DBLP:journals/corr/WangSLRWPCW14}, a common technique for improving models by first seeking the hardest examples and then emphasizing those examples through retraining or fine-tuning. Our approach notably achieves this without human labelers. See Figure~\ref{fig:scenegraphconversion} for a complete diagram of our approach and Figure~\ref{fig:intervention-example} for qualitative examples. We now explain in detail each of the components of our method.


\paragraph{The scoring function} $f$ should delineate between interventions that were minimal and those that caused a significant change in perception performance, with the assumption being that large negative (positive) changes imply that the intervention (reverse intervention) was detrimental to $\phi$. 

Our goal
in designing $f$ is to replicate the average precision (AP) score's intent, which values having few predictions with high intersection over union (IOU) to ground truth targets. Another goal was to evaluate entire scenes and not just target assets. This is important because even though our interventions can be local to a node (weather is of course global), they may still impact detecting any scene constituent. We choose not to use the popular mAP because it is defined over a dataset and thus is not suitable for identifying individual challenging scenes, which our method requires before aggregating at the intervention level. To compute $f$, we get the model's predictions and order them by descending confidence. We sequentially align each prediction with the highest IOU ground truth. If $\textrm{IOU} > .05$, an empirically chosen threshold, then we mark this ground truth as claimed. The per prediction score is the product of the prediction's confidence and its IOU. We then take the mean over all predictions to get the model's final score on this example. The result is that predictions with low confidence or poor IOU reduce the model's score, while predictions with high confidence on quality boxes increase the score.

\paragraph{Causal interventions on simulated scenes}
We draw from causal inference where interventions allow us to assess the causal links between the scene and the model's score. We change an aspect of a scene sequence $S_i$ 
, such as a rotation or location of a specific vehicle, render this new scene $S_{i}^\prime$ as $I^\prime$, and then compute the $\delta = f(\phi, I^\prime, L^\prime) - f(\phi, I, L) \in [-1, 1]$. We decide sufficiency by whether $|\delta| \ge \tau$, the aforementioned threshold parameter. After performing this procedure $N$ times, filtering by sufficiency, and then grouping by the intervention type, we arrive at a prioritized list of challenging groups defined by either rotation, vehicle type, or weather pattern.

\paragraph{Generating interventions%
}
Uniformly random interventions produce unlikely scenes under the true data distribution\footnote{Empirically, so do interventions sampled uniformly over the categories of interest.}. Even if such an intervention would identify a weakness in the detector, its utility in improving our model is unclear because such a weakness may be very far from a realistic setting. We should favor finding groups that have higher probability under the data distribution. This is especially important for a limited model capacity because learning to detect flying cars and other unrealistic low-priority scenarios might take capacity away from pressing needs.

Formally, with $p_{\textrm{R}}(x)$ as the generation process, $y$ our surrogate score, and $z$ a confounder that affects both $x$ and $y$, we need to draw a counterfactual $x^\prime$ that is independent of $z$ with which we can causally probe the model's weaknesses. Sampling from $p_{\textrm{R}}(x)$ is challenging because retrieving the same scene again with just one change is difficult. We could act directly on the scene graph and model the conditional distributions of a single node change, then select changes via Gibbs sampling, and define interventions as sampling from these conditional distributions. 
Instead, we choose to discretize the scene~\citep{oord2016wavenet,DBLP:journals/corr/EngelRRDESN17,razavi2019generating} and use masked language models~\citep{dosovitskiy2021image,khan2021transformers} because of their resounding recent success modeling distributions of combinatorial sequences relative to other approaches, as demonstrated clearly in language. Specifically, we train an MLM as a denoising autoencoder (DAE) to sample from $p_{\textrm{R}}(x)$~\citep{DBLP:journals/corr/abs-1305-6663,DBLP:journals/corr/abs-1905-12790,10.1145/1390156.1390294}, where the MLM operates on discretized scene graphs, flattened to be sequential. This provides a mechanism to sample counterfactuals from the data distribution DAE~\citep{DBLP:journals/corr/abs-2009-10195,DBLP:journals/corr/abs-1902-04094}.

For each scene drawn from the original training distribution, the MLM infers a new scene close to the original distribution by making a singular semantic change over weather, vehicle asset type, rotation, or location. For example, it may choose a vehicle instance and change that vehicle to a different vehicle type. Or it may rotate that vehicle some non-zero amount. For weather, the semantic changes could be over cloudiness, precipitation, precipitation deposits (puddles), wind intensity, or the angle of the sun (light). We never add or delete a node, only semantically change them. Because the MLM was trained to a low perplexity on data drawn from the distribution, it samples scenes that are likely under the original distribution $p_{\textrm{R}}(x)$. Because it is not the exact distribution and errors will accumulate when applying many interventions sequentially, we intervene for just one step in most of our experiments, equivalent to a single node change in the scene graph. We expand this with an investigation into what happens when we take a second successive intervention step.

\section{Related Work}
\label{sec:related-work}

\paragraph{MLM as a generator}

While we believe we are the first to propose using an MLM as a generator in order to take causal interventions, \citet{DBLP:journals/corr/abs-2009-10195} generates from an MLM in order to augment natural language task training with generated examples. \citet{DBLP:journals/corr/abs-1905-12790} and \citet{DBLP:journals/corr/abs-1902-04094} do so in order to generate high quality examples for use in downstream examples, with the former producing molecules closer to the reference conformations than traditional methods and the latter producing quality and diverse sentences. None of these operate on scene graphs.

\paragraph{AV Testing and Debugging}
See~\citet{corso2020survey} for a detailed survey on black-box safety validation techniques. We believe that we are the first to take causal interventions in static scenes to test AV detection systems, although multiple approaches~\citep{ghodsi2021generating,abeysirigoonawardena2019generating,koren2018adaptive,corso2019adaptive,o2018scalable,rempe2021generating} test AV systems through adversarial manipulation of actor trajectories and operate on the planning subsystem. ~\citet{wang2021advsim} generates adversarial scenarios for AV systems by black-box optimization of actor trajectory perturbations, simulating LiDAR sensors in perturbed real scenes. Prior research has focused on optimization techniques for adversarial scenario generation through the manipulation of trajectories of vehicles and pedestrians. They either test only the planning subsystem in an open-loop manner or the whole AV system in a closed-loop fashion. Unlike our work, they do not allow for causal factor error interpretation. We focus on open-loop evaluation of AV perception and attempt to find causal factors for performance degradation through the generation of in-distribution counterfactuals with a masked language model trained on scene graphs. Concurrently,~\citet{leclerc20213db} proposed a configurable system to diagnose vulnerabilities in perception systems through synthetic data generation. We show how to generate complex scene manipulations using the MLM and study scenes of significantly higher complexity, although it is possible in theory to implement our method within their framework.
    
\paragraph{Scene manipulation} ~\citet{ost2020neuralscenegraphs} learn neural scene graphs from real world videos via a factorized neural radiance field~\citep{mildenhall2020nerf}, while ~\citet{kar2019metasim, devaranjan2020metasim2} generate scene graphs of AV scenes that match the image-level distribution of a real AV dataset as a means to produce realistic synthetic training data. All three can be seen as a precursor to our method for handling real world data. ~\citet{dwibedi2017cut} generate synthetic training data for object detectors by learning to cut and paste real object instances on background images, which elicits a confounder because of how artificial the pasted scenes appear.
    
    
\paragraph{Adversarial detection} is another way of viewing our work. \citet{xie2017adversarial} showed that we should consider the detection task differently from the perspective of adversarial attacks, but did not explore finding root causes. \citet{DBLP:journals/corr/abs-1808-02651} use a differentiable renderer to find adverse lighting and geometry.
Consequently, images appear stitched, a confounder to the natural distribution. \citet{DBLP:journals/corr/AthalyeS17} synthesizes real 3D objects that are adversarial to 2D detectors. They are limited to single objects, moving in the location, rotation, or pixel space, and do not identify causal factors. \citet{zeng2019adversarial,tu2020physically} synthesize 3D objects for fooling AV systems, both camera and LIDAR, with a goal to demonstrate the existence of one-off examples. 



\paragraph{Challenging groups} Improving the model to recognize found groups, potentially sourced from the distribution's long tail, is an important goal. Numerous methods~\citep{ren2019learning, an2021resampling} do this by re-weighting or re-sampling the training set, with \citet{chang2021imagelevel} focusing on detection. \citet{sagawa2020distributionally} uses regularization and \citet{wang2021longtailed} uses dynamic routing and experts. All of these approaches require us to know the problematic groups in advance, which would only happen \textit{after} applying our method.
Further, they do not assess why the model is weak, but only seek to fix the problem. This makes it challenging to understand if the core issue has been addressed. \citet{gulrajani2020search} suggests that these approaches are not better than ERM, which is how we incorporate our found groups in Section~\ref{sec:experiments}.

\section{Experiments}
\label{sec:experiments}

We run a suite of experiments analyzing our method and compare it against random interventions. 


\subsection{Setup}
\label{sec:experiments-setup}

\paragraph{Model} We selected six battle-tested models from Detectron2: $18\textrm{C4}$, $18\textrm{FPN}$, $34\textrm{C4}$, $34\textrm{FPN}$, $50\textrm{C4}$, and $50\textrm{FPN}$. These are common ResNet~\citep{he2015deep} architectures that include a litany of other attributes such as Feature Pyramid Networks~\citep{lin2017feature}. We created additional configurations that are 2x, 3x, 4x, and 5x wider versions of $50\textrm{FPN}$, exemplified by $50\textrm{FPN}2\textrm{x}$, for a total of ten tested architectures. The $\textrm{C4}$ and $\textrm{FPN}$ mix provided variation in model configuration, while the $18$, $34$, and $50$ layer counts and their widths vary in parameters. We made minimal changes to account for training on our dataset and with $4$ gpus instead of $8$. All models were trained for $90000$ steps (8-9 hours) without pre-training; none reached zero training loss. 

\paragraph{Datasets} We first selected the CARLA preset map -- Town03 or Town05. Town03 is the most complex town, with a 5-lane junction, a roundabout, unevenness, a tunnel, and more. Town05 is a squared-grid town with cross junctions, a bridge, and multiple lanes per direction. Both have ample space to drive around in a scene and discover novel views. Then we randomly chose from among the pre-defined weather patterns. We sampled the camera calibration and the number $V$ of vehicle assets according to the Nuscenes~\citep{nuscenes2019} distributions, then placed those $V$ vehicles, the ego agent, and $P = 20$ pedestrian assets, at random town waypoints suitable for the asset type. Finally, we attached the calibrated camera to the ego agent and enabled autopilot for all agents. We stabilized the scene for $50$ timesteps after spawning, then recorded for $150$ steps and saved every $15$th frame. We needed the 2D ground truth boxes for each asset, but found the suggested approach\footnote{See client\_bounding\_boxes.py in the CARLA library, commit 4c8f4d5f191246802644a62453327f32972bd536.} lacking because it frequently had trouble with occlusions and other challenging scenarios. See the Appendix for heuristics we developed to help filter the ground truth boxes. For detection results on all charts, we report average precision (AP) over vehicle datasets.

\paragraph{MLM} We used the MaskedLMModel architecture\footnote{See masked\_lm.py\#L30 in the FairSeq library, commit 1bba712622b8ae4efb3eb793a8a40da386fe11d0} from the FairSeq~\citep{ott2019fairseq} library for our MLM. We train and validate on held out IID datasets of sequences converted from scene graphs, where the dataset was created as described in the prior paragraph. Encoding the scene graph language required us to translate $G$ with continuous node attributes into discrete sequence $S$. The first $10$ tokens corresponded to weather attributes (cloudiness, precipitation, sun altitude angle, etc), the next $5$ to camera intrinsics, and the following $15$ to the ego agent. After these $30$, we had a variable number of agents, each sequentially represented by $17$ tokens. The two extra tokens for the non-ego agents were related to vehicle type, which was fixed for the ego agent. Although the $10$ weather attributes were each continuous, we selected these vectors from $15$ weather choices during training and so, with regards to the encoding, they each corresponded to discrete choices. Because the camera intrinsics were drawn from the (realistic) discrete Nuscenes distribution, their encoding was also discrete.

\begin{wraptable}{r}{7.2cm}
\begin{tabular}{c|c|c}
Intervention & Percent $> 0.2$ & Total \\
\hline
\multicolumn{3}{c}{Tier 1: Likely Challenging Groups} \\
\hline
DiamondbackBike & 24.4 & 123 \\
Cloudy Dark & 19.4 & 36 \\
GazelleBike & 18.9 & 122 \\
Cloudy Dark Puddles & 17.2 & 29 \\
CrossBike & 16.5 & 121 \\
Rotation - 178 & 15 & 20 \\
Rotation - 121 & 13.0 & 23 \\
\hline
\multicolumn{3}{c}{Tier 2: Borderline Groups} \\
\hline
KawasakiBike & 6.5 & 92 \\
Cybertruck & 6.4 & 94 \\
Carla Cola & 6.0 & 198 \\
Sunny Puddles & 5.4 & 56 \\
\hline
\multicolumn{3}{c}{Tier 3: Easy Groups} \\
\hline
Citroen C3 & 1.6 & 188 \\
Mercedes CCC & 1.0 & 206 \\
\end{tabular}
\caption{Illustrative table of interventions, ordered by percent of times they were involved in a high magnitude $\delta$ edit. 
Section~\ref{sec:experiments-analysis} suggests our cutoff resides between 6.0 and 6.4.}
\label{table:intervention}
\vspace{-8mm}
\end{wraptable}

The agent tokens had a set order: discrete type (blueprint), then continuous $(x, y, z)$ locations, then $(\textrm{roll}, \textrm{yaw})$ rotations. To discretize the locations, we first subtracted their minimum possible value. The resulting $v \in [0, 600)$ was encoded with $w_0 \in [0, 5]$ for the hundreds place, $w_1 \in [0, 99]$ the ones, and $w_2 \in [0, 9]$ the decimal, so $v = 100w_0 + 10w_1 + 0.1w_2$. This small precision sacrifice marginally impacted scene reproduction. We encoded rotation similarly, albeit was bounded in $[0, 360)$.

\subsection{Interventions}


In this section, we investigate the relative ordering of groups by the MLM, where the order is determined by the degree to which that group is involved in a detrimental intervention.

Table~\ref{table:intervention} shows selected ordered results from the intervention procedure described in Section~\ref{sec:method}. We performed the procedure on $N = 10000$ test scenes $G_k$ where our $\phi$ was an $18\textrm{C}$ model trained on the base $10000$ subset from Town03 and $\tau = 0.2$. We additionally filtered the groups to those that occurred at least $20$ times in the procedure.

On the left side we see the intervention taken, for example changing a single agent type to a Cybertruck (a large truck made by Tesla) or changing the weather such that it is now sunny with reflective puddles.
The second column shows the percentage of scenes that the intervention produced a $\delta \ge 0.2$. We include both when the change was \textit{to} that target and the delta was negative as well as when it was \textit{from} that target and the delta was positive. The last column in the table reports how many times in total this intervention occurred in the $10000$ scenes.

Summarizing the table, we find that a handful of asset switches appear to be detrimental for the model according to this metric. Small bikes had an outsized effect, as did cloudy weather and the rotations where a car faced the ego agent or turned to the left. Just after the last bike are two large vehicles, the Cybertruck and the Cola Car. The specificity of the weathers and rotations are because they are translations of our discretization. Practically, there is a range of rotation and weather values around the group that would all suffice. Finally, we do not include location results in the table because the MLM frequently re-positioned the asset outside the camera's view. This said more about the asset than it did about the location and was rife with confounders based on what was behind that asset. 
We could have localized the location interventions more by masking MLM options, but leave that for future work. 

\begin{wrapfigure}{l}{0.48\textwidth}
  \begin{center}
    \includegraphics[width=0.47\textwidth]{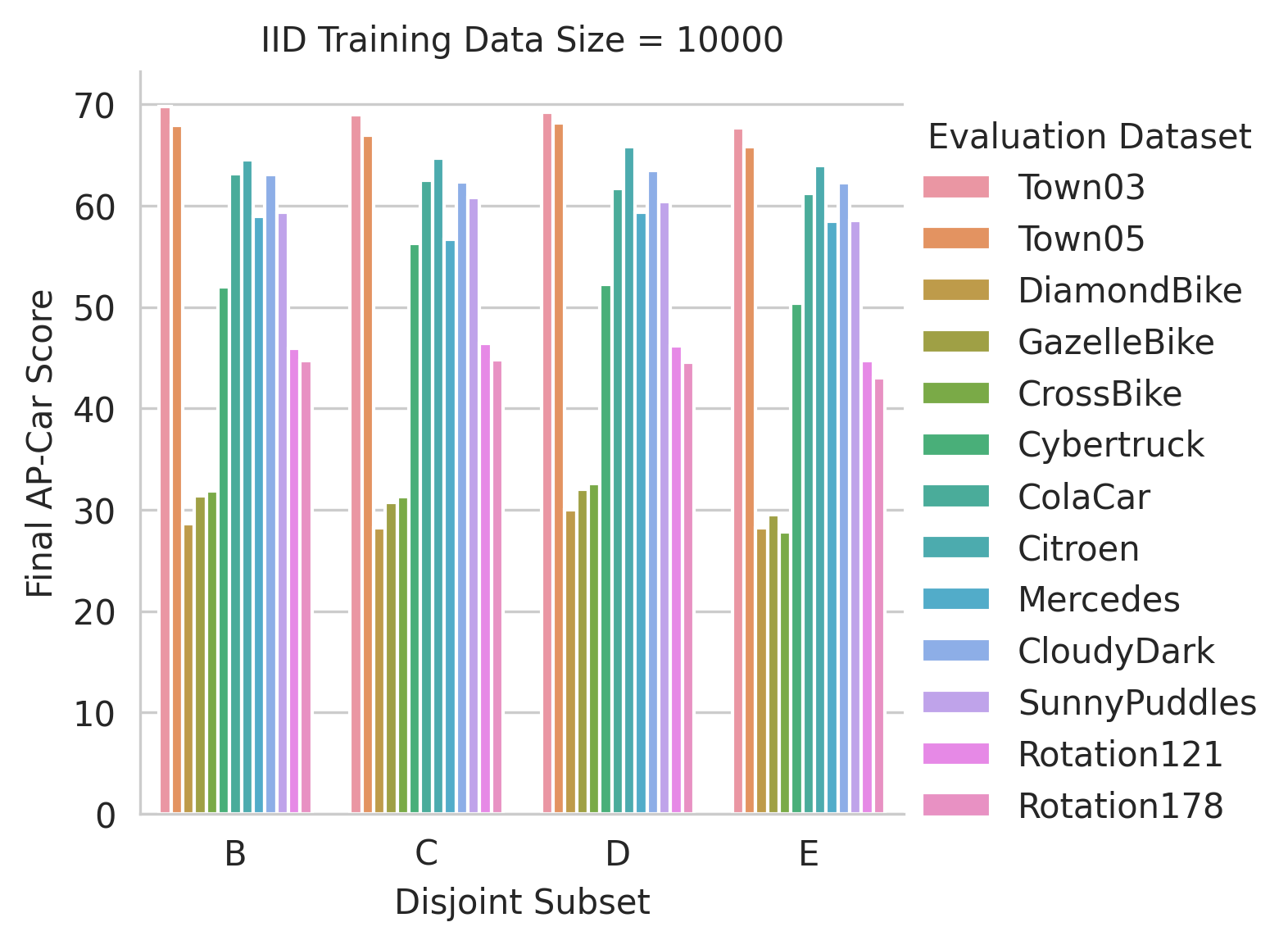}
  \end{center}
    \caption{Test results with config $18\textrm{C}4$ when training on disjoint IID subsets. Results are consistent, suggesting that the harder groups - bikes, rotations, and cybertruck - are ubiquitously hard.}
    \label{fig:exp7-18c4}
\end{wrapfigure}

\subsection{Analysis}
\label{sec:experiments-analysis}

After obtaining candidate groups from the designed interventions, we investigated the effect of modifying the data sampling procedure to increase the prevalence of these groups by building and evaluating datasets sampled from the MLM training set. For asset groups, for each datum, we uniformly sampled $n_{\textrm{v}} \in [3, 6]$ vehicles selected from the scene. We then randomly chose vehicles $v_0, v_1, \ldots, v_{n_{\textrm{v}}}$ in that scene, including vehicles that may not be in the camera's purview, and changed them to be the target group. So as to not accidentally introduce a bias through the random process, we selected the same vehicles $v_k$ for all group datasets.
For rotation groups, we chose those same vehicles but rotated them to be the target rotation instead of switching their asset. For weather groups, we changed those scenes to have the target weather instead.

\paragraph{Does our method correlate with AP score?} Figure~\ref{fig:exp7-18c4} shows evaluation results on these groups when training $18\textrm{C}4$ on four disjoint $10000$ sized subsets of the data. The models performed best on the IID data from Town03 and just a little bit worse on the same from Town05. Further, they did exceptionally well on those two datasets, validating that they were trained sufficiently. The group results are mostly in line with our expectations from the interventions - the models did well on Citroen and Mercedes, poorly on the rotations, and terribly on the bikes. There is a large jump from the reasonable results on ColaCar and SunnyPuddles to the mediocre results on Cybertruck, which is directionally correct per Table~\ref{table:intervention}. However, the strong results on CloudyDark are surprising.

\begin{figure}[t!]
    \centering
    \includegraphics[width=0.70\textwidth]{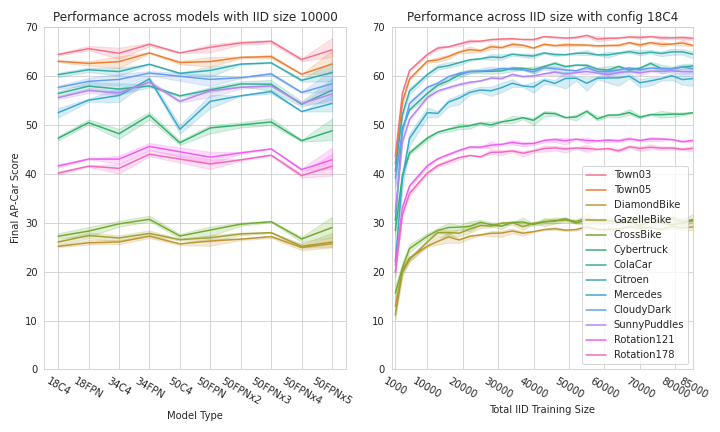}
  \caption{Independently increasing the model capacity (left) and increasing the data size (right). No model distinguished themselves and we quickly taper in how effectively the model utilizes the data. We consider the dip in the capacity chart to be an artifact of the training procedure and using the same settings for all models.}
  \label{fig:exp1and2-18-10k}
\end{figure}


Summarizing, if the threshold for choosing a group is between $5.5\%$ and $6.5\%$ and we focus on interventions affecting vehicles directly (rotation and type), then our method correlates well with empirical results. We have likely not found the exact causes plaguing the model, but we have narrowed them greatly. The model's regression when changing a car to a bike may be because it performed poorly on bikes. It may also be because the car was occluding another vehicle or that it itself was not occluded. This is especially true in light of the weather results suggesting that weather is not a conclusive factor. Finding the exact cause is difficult, even in simple settings~\citep{arjovsky2020invariant}. We leave such improvements for future work.

\begin{figure}[!t]
    \centering
    \includegraphics[width=\textwidth]{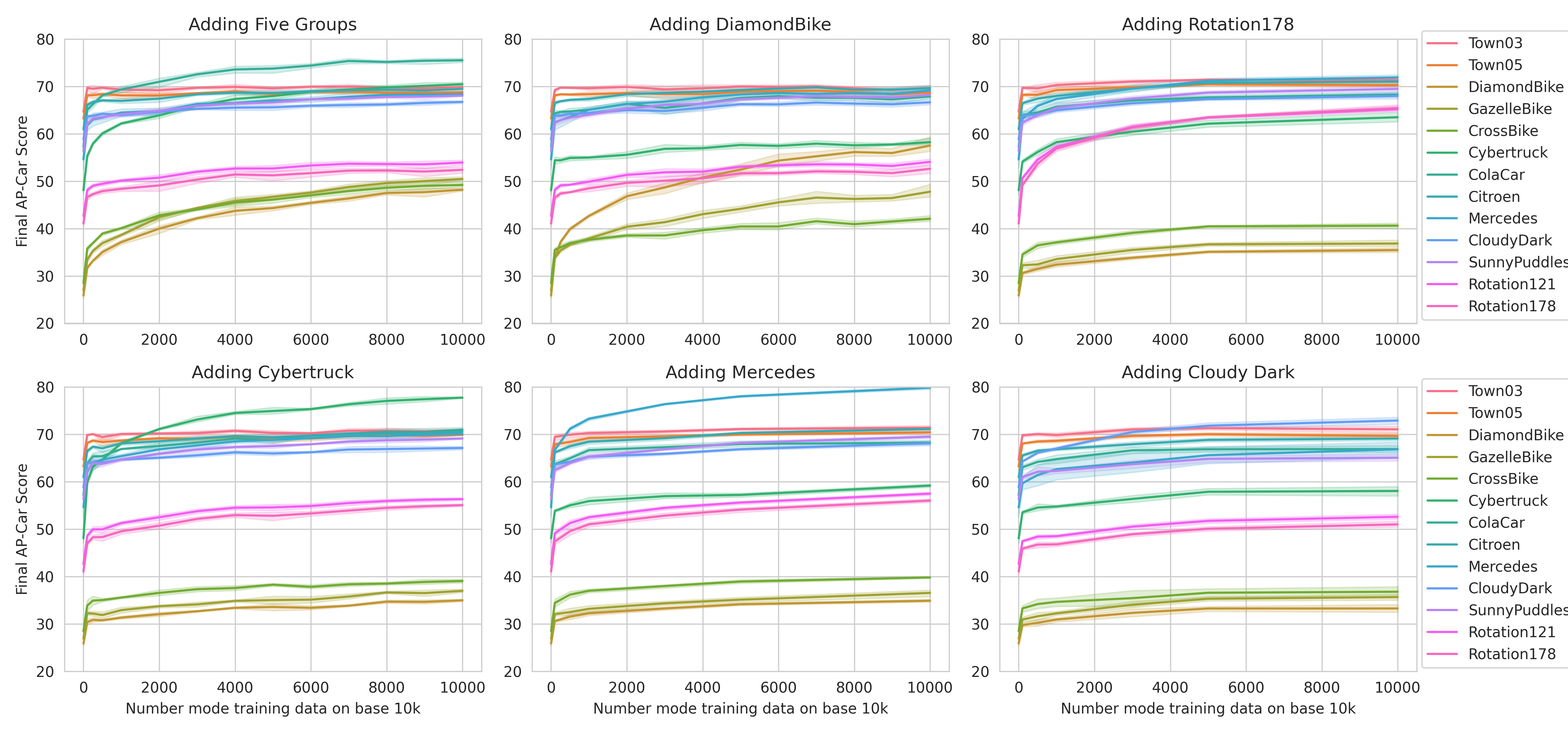}
    \caption{Results of training $18\textrm{C}4$ on the base IID $10000$ training set plus additional group data. The five groups in the top left (Cybertruck, Cola Car, Diamondback, Gazelle, and Crossbike) were added equally. For all charts, adding any one group improved all of the other evaluation scores, and at no point did we lose efficacy on the IID data as a whole. Figure~\ref{fig:exp3-configs-0-1k} (Appendix) zooms in on the initial jump.}
    \label{fig:exp3-configs-0-20k}%
\end{figure}

\paragraph{Can we address these issues by increasing capacity?} Recent papers~\citep{zhai2021scaling,DBLP:journals/corr/abs-2102-06701} suggest that scaling our models will improve results. An affirmative answer would mean we would not need to collect more data. The left side of Figure~\ref{fig:exp1and2-18-10k} suggests a negative answer when testing over a range of model sizes and types.  
We see that no model was distinguished with respect to their final values when training on $10000$ IID examples.

\paragraph{What if we increased IID data?} This is preferable because IID data is easier to collect than group specific data. The right side of Figure~\ref{fig:exp1and2-18-10k} suggests this will not be sufficient. Test efficacy on town and group data jumped from $1000$ to $10000$ IID examples, but then slowed precipitously. Figure~\ref{fig:exp.3&4-18_C4} (Appendix) affirms that this is unlikely to change by suggesting that the percentage of representation of the group is what matters, rather than absolute count.

\begin{wrapfigure}{l}{0.53\textwidth}
  \begin{center}
    \includegraphics[width=0.52\textwidth]{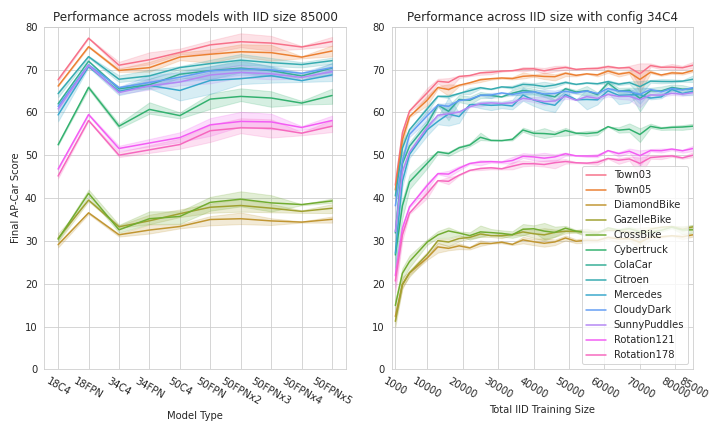}
    \end{center}
    \caption{Increasing both data and model capacity at the same time. The left side ranges over model capacity with maximum IID data size ($85000$), while the right side ranges over IID data size with a bigger model - $34\textrm{C}4$.}
    \label{fig:exp1and2-34-85k}
\end{wrapfigure}

\paragraph{What if we increase data and capacity simultaneously?} Results remained negative, as seen in Figures~\ref{fig:exp1and2-34-85k} and \ref{fig:bothoverlay} (Appendix). The left graphic in Figure~\ref{fig:exp1and2-34-85k} evaluates all models on $85000$ examples and the right one shows results for just the $34\textrm{C}4$ model across a range of IID data counts. First, observe that all of the models have similar evaluation scores. Second, they all struggled on the harder groups. And third, as seen more clearly in Figure~\ref{fig:bothoverlay}, more data yielded a small accretive effect. All else equal, adding data may be better than adding model capacity. 


\begin{figure}[!t]
    \centering
    \includegraphics[width=0.8\textwidth]{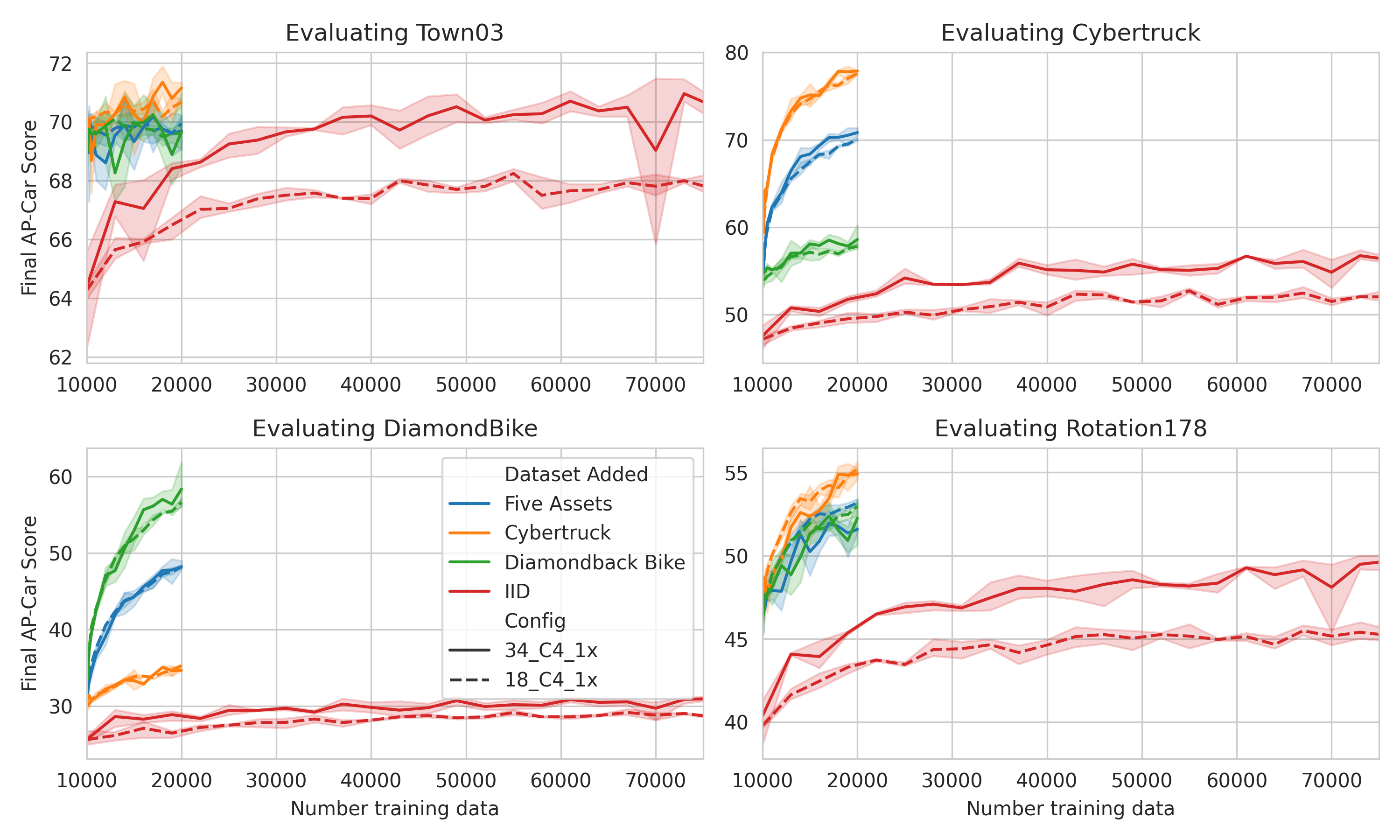}
    \setlength{\belowcaptionskip}{-10pt}
    \caption{How much IID data is required to match a small amount of extra hard group data. Top left shows $20000$ more IID data was required to reach par on IID with $250$ group data. Bottom left shows that we never reached the same level on Diamondbacks with IID data as with adding Cybertrucks, let alone actual bikes.}
    \label{fig:exp23-iidmodes}
\end{figure}

\paragraph{Using group data} We expect that adding data from the groups to the training set will improve performance on that group. The top left plot in Figure~\ref{fig:exp3-configs-0-20k} confirms this. We added an even amount of each group to the base $10000$ IID subset and see that every group improved without impacting the Town03 and Town05 results. The other plots in Figure~\ref{fig:exp3-configs-0-20k} show what happens when we add in training data from any one group $M$. This predictably improved the model's results on $M$'s validation set. It surprisingly also improved results on \textit{all} of the other $M^\prime$ and the Town data. The improvement to $M^\prime$ is smaller than that to $M$, but it is notable. The gains for a specific group were more pronounced for like groups - adding data from a biker group (Diamondback, Omafiets, Crossbike) improved the other biker groups more than adding data from the heavy car groups (Cybertruck, Colacar), and vice versa. Adding rotation groups helped ubiquitously albeit not as much as adding a bike group did for the other bikes. The least effective fix was adding the CloudyDark weather mode. Figure~\ref{fig:exp.3&4-18_C4} shows that these trends persisted for a base of $85000$ IID data as well. 

\paragraph{Comparison with random interventions}

As we alluded to in Section~\ref{sec:method}, taking random interventions is problematic because whether the group is reasonable for the distribution will be a confounder. We wish to prioritize the found groups to be those that are more likely seen in the wild. We show here that this is true by taking the $10000$ source scenes used for the MLM interventions and applying random manipulations of the same type. For example, if we changed agent $a_j$'s vehicle type in $G_k \rightarrow G^{\textrm{MLM}}_k$, then we changed $a_j$ to a random vehicle type in $G_k \rightarrow G^{\textrm{Random}}_k$.


Table~\ref{table:random} shows results for random and MLM interventions over the same assets from Table~\ref{table:intervention}. Observe that the assets were ordered incorrectly with CarlaCola higher than both Cybertruck and Kawasaki Bike. Random also had a higher percent of high magnitude threshold events; In general, $13.2\%$ of random interventions impacted the model versus $10.2\%$ of MLM interventions. We hypothesize this is because random re-sampling of elements of the scene graphs corresponded to sampling from a data distribution that does not faithfully represent the original training distribution. A $3\%$ difference is large with respect to how much extra work would be required by humans combing through the data for plausibility and whether to include in retraining.

\begin{table}[b!]
\centering
\begin{tabular}{c|cc|c}
Intervention & MLM & Rand & Rand Total \\
\hline
CrossBike & 16.5 & 19.0 & 126 \\
GazelleBike & 18.9 & 18.7 & 171 \\
DiamondbackBike & 24.4 & 15.8 & 152 \\
\hline
Carla Cola & 6.0 & 8.1 & 210 \\
Cybertruck & 6.4 & 6.8 & 176 \\
KawasakiBike & 6.5 & 6.6 & 121 \\
\hline
Citroen C3 & 1.6 & 3.5 & 197 \\
Mercedes CCC & 1.0 & 3.8 & 183 \\
\end{tabular}
\caption{Results for MLM and Random asset intervention strategies, ordered by the percent of times that they were involved in a high magnitude $\delta$ \textit{random} event. While the top three are the same, Random flubbed the dividing line by placing a) Cybertruck above Kawasaki and b) Carla Cola well ahead of both. Its failure rate for the easy cars was much higher and, in general, posited 3\% more failures than MLM. All told, its results created more need for human verification and testing and reduced the degree of automation that we could employ to find hard groups.}
\label{table:random}
\end{table}

\begin{wrapfigure}{l}{0.48\textwidth}
  \begin{center}
    \includegraphics[width=.47\textwidth]{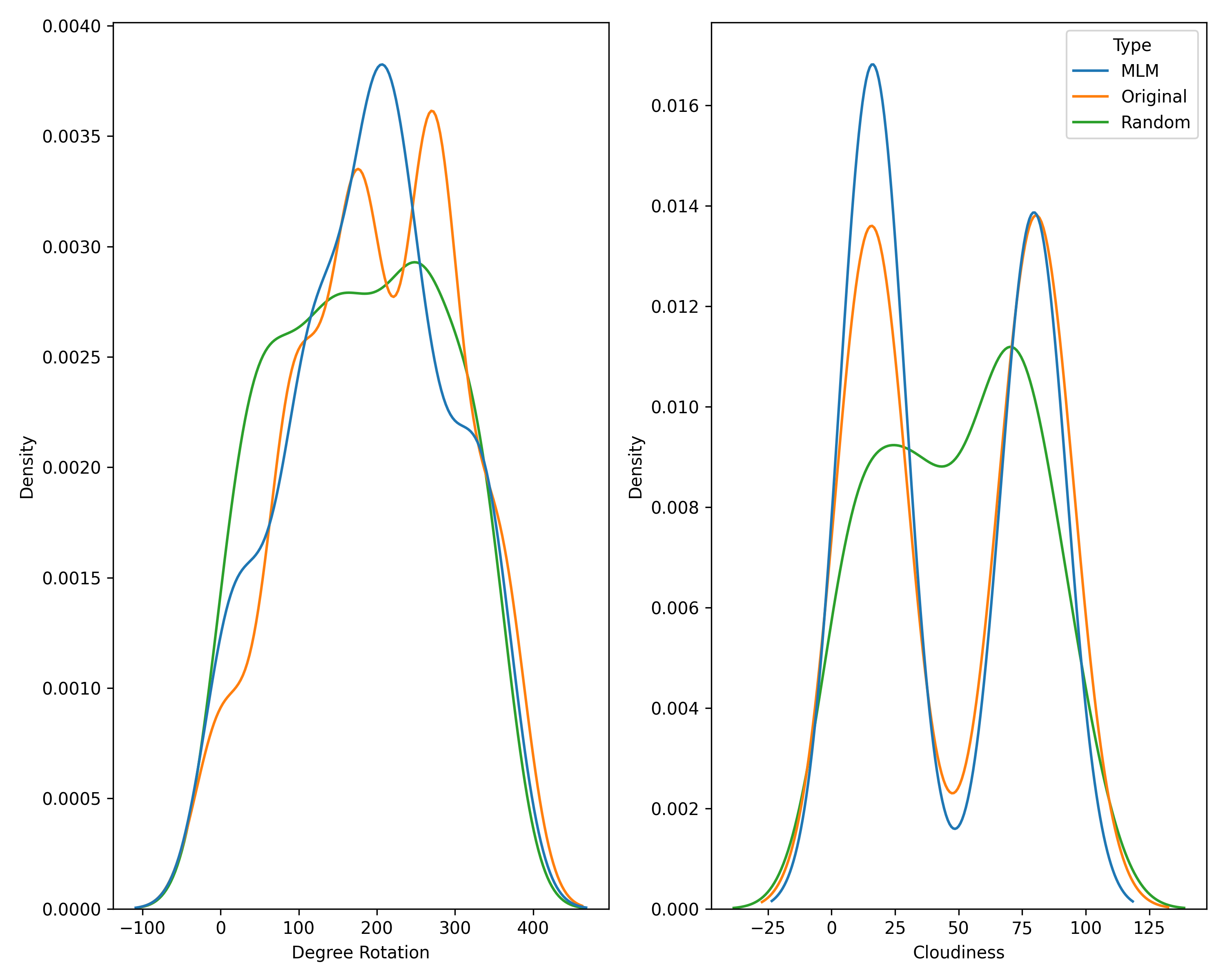}
    \end{center}
    \caption{Comparing rotation and weather results for MLM and Random intervention strategies. MLM aligns with Original much better than Random does. Further, Random has a much wider berth of possible problematic modes, a concern given practical limits to model capacity and data budgets.}
    \label{fig:intervention_kde}%
\end{wrapfigure}

Figure~\ref{fig:intervention_kde} shows density plots for rotation and cloudiness interventions, conditioned on the intervention having been detrimental. We use density plots to demonstrate the differences between Random and MLM because these interventions are continuous for Random. For rotation, there was a mostly steady plateau for Random while MLM showed a clear single group aligned with the bi-modal humps in Original. For weather, Original and MLM were almost overlapping and, while Random was similarly bi-modal, its shape was less pronounced and more even as expected. These both reinforce our claim that the advantage of MLM is that it gears us towards higher priority groups to fix that are in line with the actual data distribution.

\paragraph{Comparison with cause-agnostic data collection} We saw in Figures~\ref{fig:exp3-configs-0-20k} and \ref{fig:exp.3&4-18_C4} (Appendix) that adding group data into training not only addresses the issue for that group but even improves the performance on other groups. The cost is that we have to perform the entire described procedure to find our interventions and \textit{then} cast a net for data of those types in order to retrain the model. An important baseline comparison would be to find data instances where the model performs poorly on the aforementioned scoring function (Section~\ref{sec:method}) and retrain by including those alongside IID data. This approach, which we christen cause-agnostic data collection, would save us the need to take interventions or gather type-specific data to retrain.

\begin{figure}[!t]
    \centering
    \includegraphics[width=\textwidth]{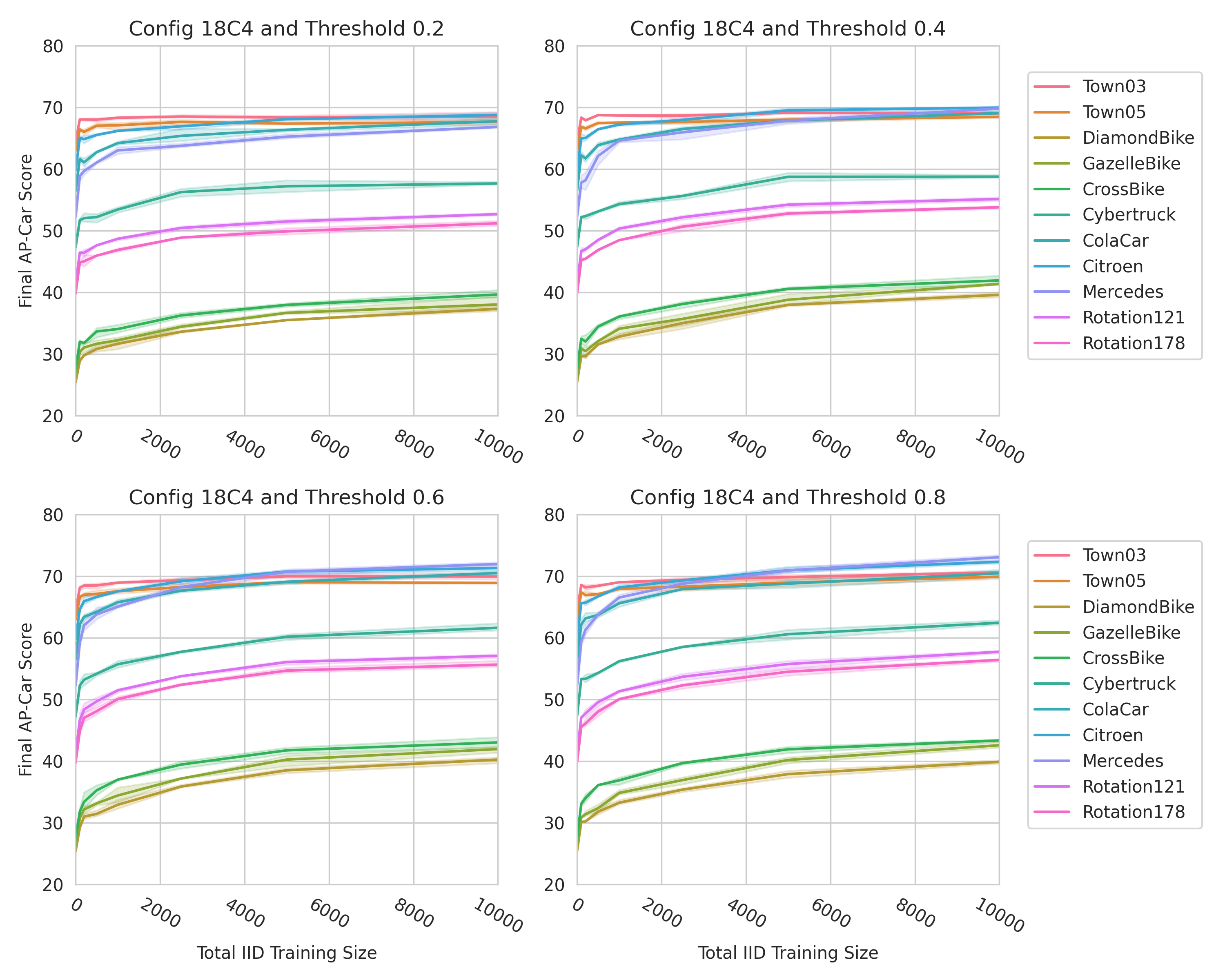}
    \caption{\textbf{Baseline cause-agnostic data collection} results. We train $18\textrm{C}4$ on the original IID $10000$ training set plus additional cause-agnostic data. The latter is chosen by first selecting a threshold from $[0.2, 0.4, 0.6, 0.8]$, then randomly selecting simulated data for which the model gets at most that score using our scoring function from Section~\ref{sec:method}. The graphs suggest a slight AP increase as the threshold increases, likely because lower threshold scores lean disproportionately towards fewer annotations and emptier scenes. Comparing these results with Figure~\ref{fig:exp3-configs-0-20k}, we see that this baseline is comparable for arbitrary groups, like the Rotations, but unsurprisingly much worse for data-specific improvements. For example, the first and second charts of Figure~\ref{fig:exp3-configs-0-20k} show that our method achieves much higher gains in the bike classes.}
    \label{fig:newexp1-18C}
\end{figure}

Figures~\ref{fig:newexp1-18C} and \ref{fig:newexp1-34C} (Appendix) show grids of results with this approach, respectively for each of our two configurations, covering four threshold values - $0.2$, $0.4$, $0.6$, and $0.8$\footnote{We did not include results on thresholds below $0.2$ because they showed the same trend.}. We test all thresholds because we do not know which will be best a priori. We then randomly draw $150000$ IID scenes, test on these scenes, and filter into buckets based on whether the resulting score was less than the given threshold. We randomly choose $10000$ scenes from each bucket and add them in tranches to the original $10000$ IID data training set. 

Observe first that the model's performance increases across the board with this data. For example, on the bikes, which were the most challenging groups, the model increases from below 30 to hover around 40 as more data is added. Next, as expected, the $34\textrm{C}4$ model is a bit better than the $18\textrm{C}4$ model for all thresholds. Third, as the threshold increases, the results improve. One hypothesis why is because the lower threshold datasets have fewer annotations and consequently emptier scenes than the higher threshold datasets.

Most importantly, how does this compare to our proposed approach? The best results for this baseline are found in threshold $0.8$. Compared against the first chart in Figure~\ref{fig:exp3-configs-0-20k} - `Adding Five Groups' - we see that the IID Town03 and Town05 results are about the same, the easier classes (Mercedes and Citroen) slightly surpass our strong results, and the Rotation results are better than ours (high 50s versus low 50s). However, for the classes where we actually add data, our method's results are much better than the cause agnostic results. For example, the most challenging groups - the bikes - reach only an AP score of 43 with cause-agnostic collection but go above 50 with our method. This is not surprising as adding group-specific data \textit{should} boost the performance. In this light, our method's advantages over this baseline are clear. First, we can ascertain which of the groups are actually problematic. This is no small feat; without our method, we would not have actually known which groups to track when performing cause-agnostic data collection. And second, we still produce a large gain over cause-agnostic data collection when we add in group-specific data. That this effect is even more pronounced for the challenging groups suggests that our method is integral for understanding on \textit{which} groups we should spend the additional capital necessary to produce representative datasets.

\paragraph{Why do these groups exist?} With causal groups in hand, we can ascertain why our models failed: The bikes are underrepresented in Nuscenes; The model rarely saw turning cars (Rotation 121) due to the town layout; The model rarely saw cars facing it (Rotation 178) due to the traffic policy and car quantity; The large cars cause occlusion labeling issues, Cybertruck more so than Cola car. Without the groups, these issues can only be hypothesized.

\phantomsection
\label{sec:secondstep}
\paragraph{What happens if we take another step?} We analyze what happens when we take a successive intervention step with the MLM to refine our causal understanding. We consider the following, where $\delta_{kj} = f(\phi, I_j, L_j) - f(\phi, I_k, L_k)$, the change in the model's efficacy from when it evaluates scene $k$ to when it evaluates scene $j$.
\begin{enumerate}
    \item Which second steps are detrimental to the one-step edited scene with threshold of $\tau_2 = 0.2$? This assesses which refinements are impactful to first edits that have a minor effect. Here, $\delta_{10} \geq \tau_1 = 0.2$ and $\delta_{21} \geq \tau_2 = 0.2$, which together imply that $0.8 \geq \delta_{10}$ because all $\delta < 1$.
    \item Which pairs are detrimental to the original scene with a threshold of $\tau_2 = 0.2$, regardless of the first step's result? This is assessing which pair of refinements are most worth exploring. Here, $\delta_{20} \geq \tau_2 = 0.2$.
    \item Conditioned on the one-step scene passing a threshold of $\tau_1 = 0.2$, which two-step scenes are as bad, i.e. they pass a threshold of $\tau_2 = 0.0$\footnote{We do not report where strictly $\tau_2 > 0.0$ because those results disproportionately tilt towards scenarios where the first edit was only somewhat detrimental. This is because if the first edit brought the score to $\tau_2$ or less, then the second edit would not pass muster regardless of its efficacy.}? Here, $\delta_{21} \geq 0$ and $\delta_{10} \geq \tau_1 = 0.2$.
\end{enumerate}

So that the search space is not prohibitively large, we limit the possible first step we take to be uniformly randomly chosen from a set $J$ that we previously analyzed and which represent a wide cross section of the challenging interventions - $J = \big\{$Diamondback Bike, Gazelle Bike, Crossbike, Cybertruck, Carla Cola, Cloudy Dark (CD), Sunny Puddles (SP), Rotation 178, Rotation 121$\big\}$. We further limit the second step to be from a different category than the first, e.g. if the first choice was an asset change, then the second step must be either a random rotation or weather change. This second step is performed similarly to how we did the original interventions, albeit $N = 60000$ times instead of $10000$ . After producing these scenes, we then score them on the same $18\textrm{C}4$ model trained on the base $10000$ subset from Town03. 

\begin{table}[t!]
\centering
\begin{tabular}{c|c|c}
Intervention & Percent $> \tau_{1, 2}$ & Total \\
\hline
\multicolumn{3}{c}{Question 1} \\
\hline
GazelleBike & 21.4 & 229 \\
CrossBike & 20.2 & 282 \\
Carla Cola & 18.9 & 355 \\
DiamondbackBike & 18.3 & 218 \\
BMW Isetta & 16.9 & 438 \\
KawasakiBike & 16.3 & 282 \\
Harley Davidson & 16.2 & 321 \\
Yamaha YZF & 15.6 & 257 \\
Rotation 10 & 13.2 & 120 \\
\hline
\multicolumn{3}{c}{Question 2} \\
\hline
(CD, KawasakiBike) & 62.5 & 72 \\
(SP, Harley Davidson) & 61.5 & 91 \\
(SP, Rotation 30.0) & 60.4 & 53 \\
(SP, KawasakiBike) & 60.3 & 73 \\
(SP, Isetta) & 58.7 & 121 \\
(SP, Grand Tourer) & 56.6 & 129 \\
(CD, Harley Davidson) & 54.3 & 92 \\
\multicolumn{3}{c}{\ldots First non-weather starting edit at 113 \ldots} \\
(Cybertruck, Rotation 160) & 41.3 & 63 \\
\hline
\multicolumn{3}{c}{Question 3} \\
\hline
(SP, Rotation 162) & 20.9 & 67 \\
(Carla Cola, Rotation 79) & 20.8 & 48 \\
(CD, Volkswagen) & 20.8 & 125 \\
(CD, CrossBike) & 20.3 & 64 \\
(CD, Rotation 290) & 19.8 & 81 \\
(SP, Jeep) & 19.4 & 103 \\
(CD, DiamondBike) & 19.3 & 57\\
(CD, Patrol) & 17.8 & 107 \\
\hline
\end{tabular}
\caption{Illustrative table of second-step interventions, ordered by the percent of time that they were involved in a high magnitude $\delta$ intervention. See the `What happens if we take another step?' paragraph in Section~\ref{sec:secondstep} for analysis.}
\label{table:twostep}
\end{table}

Results in Table~\ref{table:twostep} address each question. For Question 1, the small vehicles are again the most problematic interventions, with four bikes, the Isetta (small car), and the two motorcycles (Harley and Yamaha) all in the top eight. After Rotation $10$, which is a new addition, there are no second edits for which at least $9\%$ pass the threshold. Because this question requires that the first intervention was not (too) detrimental - otherwise the second intervention would not be able to pass the $\tau_2 = 0.2$ threshold - that these results are similar to the prior results in Table~\ref{table:intervention} is not surprising.

For Question 2, we see very high probability detrimental pairs. Additionally, the first time a non-weather appears as the initial intervention is not until index 113. That the weathers are appearing first is explainable by there being only two weather options possible in the first intervention (by fiat), which makes it easier for them to be selected first than asset changes. There are many more weathers possible in the second intervention, and so any one of them has a hard time distinguishing itself, which makes it challenging for a (rotation, weather) or (asset, weather) pair to appear. 

\begin{wrapfigure}{l}{0.48\textwidth}
  \begin{center}
    \includegraphics[width=.47\textwidth]{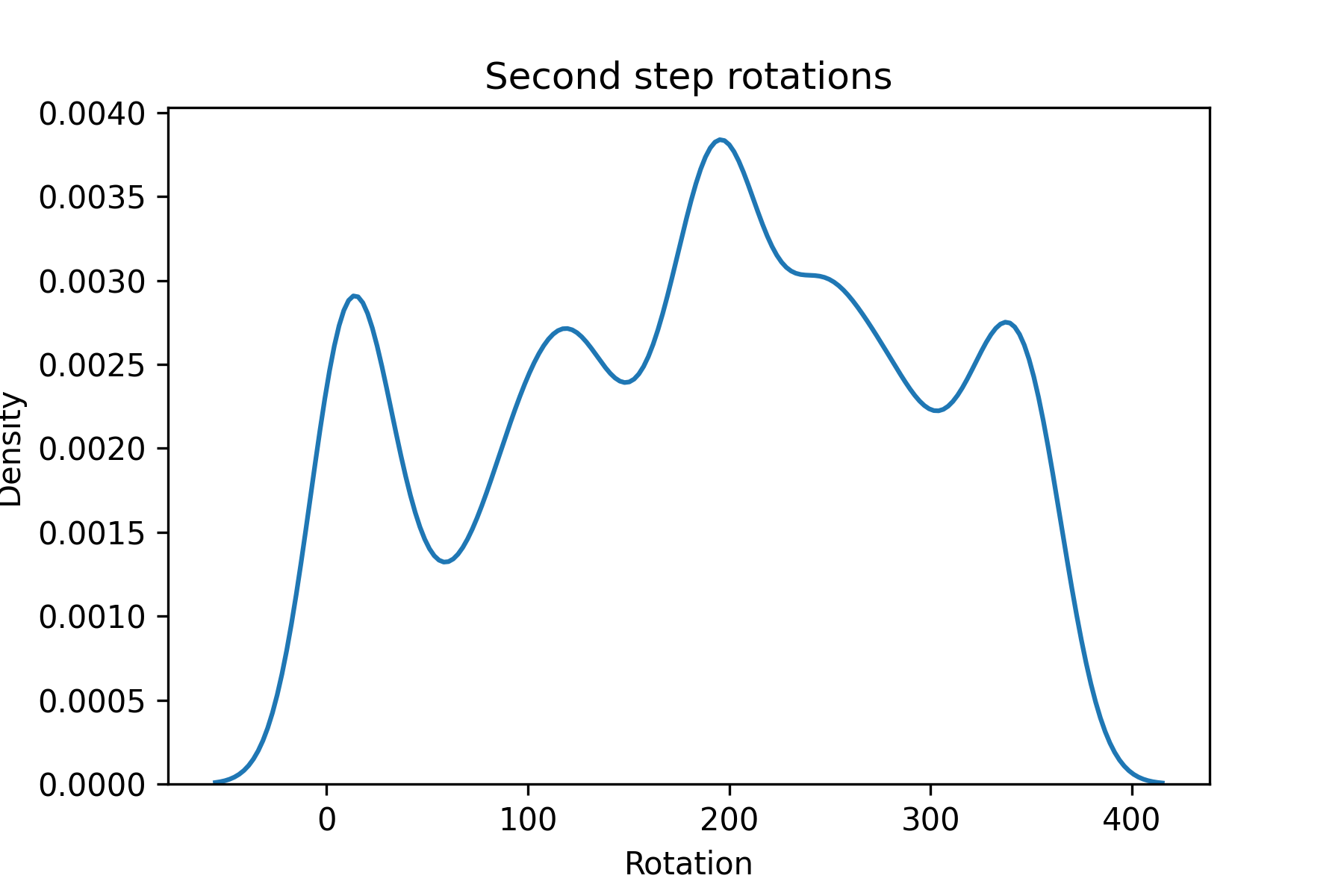}
    \end{center}
    \caption{Rotation density plot for second interventions, conditioned on the intervention being detrimental. That the shape of this plot is very different from the MLM and Original plots in Figure~\ref{fig:intervention_kde} suggests that the MLM is applying a different distribution on the second intervention. In other words, it has already drifted.}
    \label{fig:secondstep_rotation_kde}%
\end{wrapfigure}

However, we are not actually sure why the probabilities are so high. They suggest that it is quite easy for a pair of interventions to confuse the model. Figure~\ref{fig:secondstep_rotation_kde} suggests that the MLM is already off of the data manifold given that the second-step rotations it is choosing have such a different distribution than the selections we see in Figure~\ref{fig:intervention_kde}. That being said, it is surprising to us that making the weather sunnier and then changing an asset to a bike for example has such a detrimental effect.

Question 3 is asking which second interventions do not improve the score given that the first intervention was sufficient detrimental. We see a high concentration of first-step weathers in the top, but it is not as ubiquitous as it was in Question 2. While not shown, the results continue to have higher than $10\%$ probabilities up to place $113$, with an asset change usually mixed in in at least one intervention.

\section{Conclusion}

Combining causal interventions, MLMs, and simulation, we presented a novel method that finds challenging groups for a detection model in foresight by having the MLM resample scene constituents. These interventions help identify and prioritize groups with poor performance without humans in the loop. We demonstrate our advantage against a baseline using cause-agnostic data upon which the model performs poorly. Our approach is a significant step towards addressing safety-critical concerns in AV. 
Beyond AV, we think the associated will benefit the causality community because the current state of the art~\citep{DBLP:journals/corr/abs-2012-07421} involves static datasets with low complexity tasks. 

Our method has limitations. We cannot yet apply it to real world data because we need full control over the scenes for the MLM to properly operate.
\citet{ost2020neuralscenegraphs} is a step towards overcoming this concern. Until then, the so-called sim2real gap~\citep{sadeghi2017cad2rl,10.1007/3-540-64957-3_63} is ever-present. 
Another limitation is that while we do show compelling results when taking a second step, these results also suggest that the MLM is already drifting from the data distribution and so its utility is reduced. In light of this, we do not expect our method to continue to work for many steps without further research because the samples will inevitably drift from the data distribution. Intervening multiple times is necessary for understanding complicated causal interactions. Each of these two limitations are of course also potential future directions. A final one is understanding better why many groups improved when adding a single group, which remains a compelling question.


\clearpage
{\small
\bibliographystyle{plainnat}
\bibliography{refs}
}

\clearpage

\appendix

\section{Appendix}

\newif\iftabon
\tabontrue
\iftabon
\subsection{Table of notation}

\begin{table}[!h]
    \centering
    \begin{tabular}{c|l}
        Symbol & Description\\\hline
        $\phi$ & Detection model \\
        $x$ & Scene \\
        $G$ & Scene graph \\
        $d \in \mathbb{N}$ & Number of objects in scene\\
        $S \in \mathbb{N}^O(d)$ & Sequence encoding of scene graph\\
        $I \in \mathbb{R}^3$ & Scene image \\
        $L \in \mathbb{R}^{4\times d}$ & Scene bounding box labels \\
        $l_k \in \mathbb{R}^4, k < d$ & The bounding box of the $k$th object \\
        $R$ & Scene generation process\\
        $p_R(x)$ & Distribution over scenes\\
        $f : (\phi, I, L) \rightarrow \mathbb{R}$ & Per-example scoring function \\
        $\delta \in \mathbb{R}$ & The change in score by intervention: $\delta = f(\phi, I^\prime, L^\prime) - f(\phi, I, L)$\\
        $\tau \in \mathbb{R}$ & Threshold value for classifying interventions as detrimental
    \end{tabular}
    \caption{Table of notation}
    \label{tab:notation}
\end{table}
\fi

\subsection{Dataset details}

CARLA does not spawn agents that collide with the environment, even the ground. To ensure agents are grounded, for any agent spawn collision, we increase its Z coordinate and try respawning. This allows us to place every agent on the map, albeit some of the conflicting agents have to `drop' from above, and consequently we wait for $50$ timesteps so those agents can settle. In that duration, the autopilot policy guides the agents to satisfactory positions. After those $50$ steps, we then record for another $150$ steps and save every $15$th frame. The resulting episodes each have ten frames with an initial distribution influenced by Nuscenes and CARLA, and a traffic policy influenced by CARLA.

We found the existing suggested approach for getting 2D ground truth boxes lacking because it frequently has trouble with occlusions and other challenging scenarios, so we developed the following heuristics to help filter the boxes. While not airtight, the resulting ground truths were qualitatively more reliable.

\begin{itemize}
    \item Filter Height: We require that the final $2d$ box is at least $30$ pixels. This is in between the easy ($40$) and medium/hard ($25$) settings on KITTI~\cite{Geiger2012CVPR}.
    \item Max Distance: We require that the ground truth detection not be more than $250$ meters away. We enforce this through the use of a depth camera attached to the ego agent. 
    \item Visible Pixel Percent (VPP) and Min Visible Count (MVC): The $2$D box is attained by pairing the $3$D box with the camera's calibration. With the latter, we get the closest point $P$ to the ego agent. We then get the depth camera's output at the $2$D box. VPP asks what percent $t$ of that box is closer than $P$ and filters it if $t \ge 80$, ensuring that at least $20\%$ of the object is not occluded. MVC asks how many pixels $q$ are further than $P$ and filters it if $q < 1300$, ensuring that the occluded object is big enough. 
\end{itemize}



\clearpage

\subsection{Supporting charts}

\begin{figure}[!h]
    \centering
    \includegraphics[width=\textwidth]{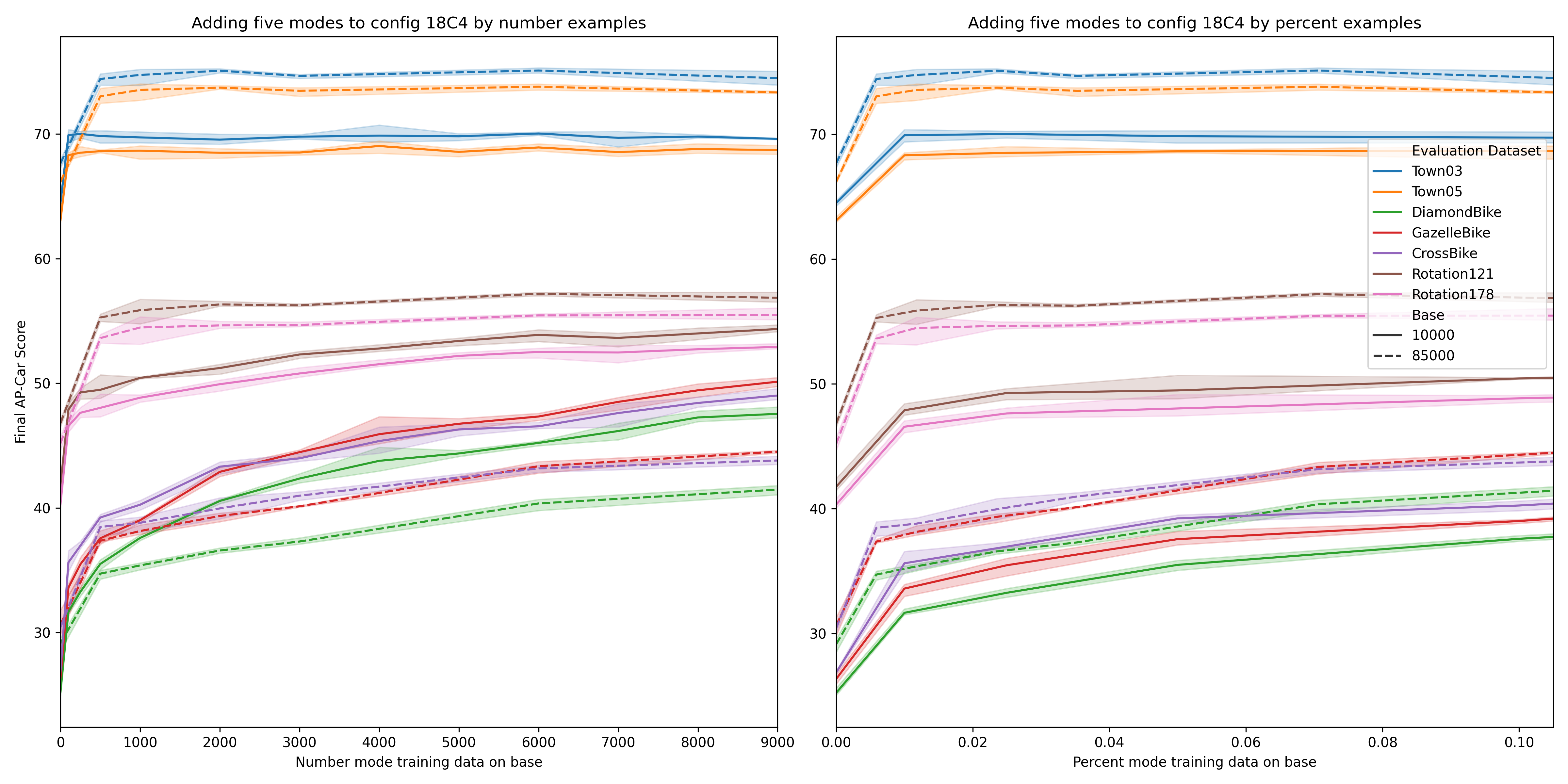}
    \caption{Performance of $18\textrm{C}4$ on select test sets when adding mode data from the three bikes, the ColaCar, and the Cybertruck on top of either $10000$ or $85000$ base IID data. Towards improving the results, these two charts show that it is not the absolute count of the mode data that is important but rather the percent of it relative to the IID data. We see that in how the trendlines for the two bases are only consistent in the percent chart. The other modes are not shown for clarity but it holds in general.}
    \label{fig:exp.3&4-18_C4}%
\end{figure}

\begin{figure}[!h]
    \centering
    \includegraphics[width=\textwidth]{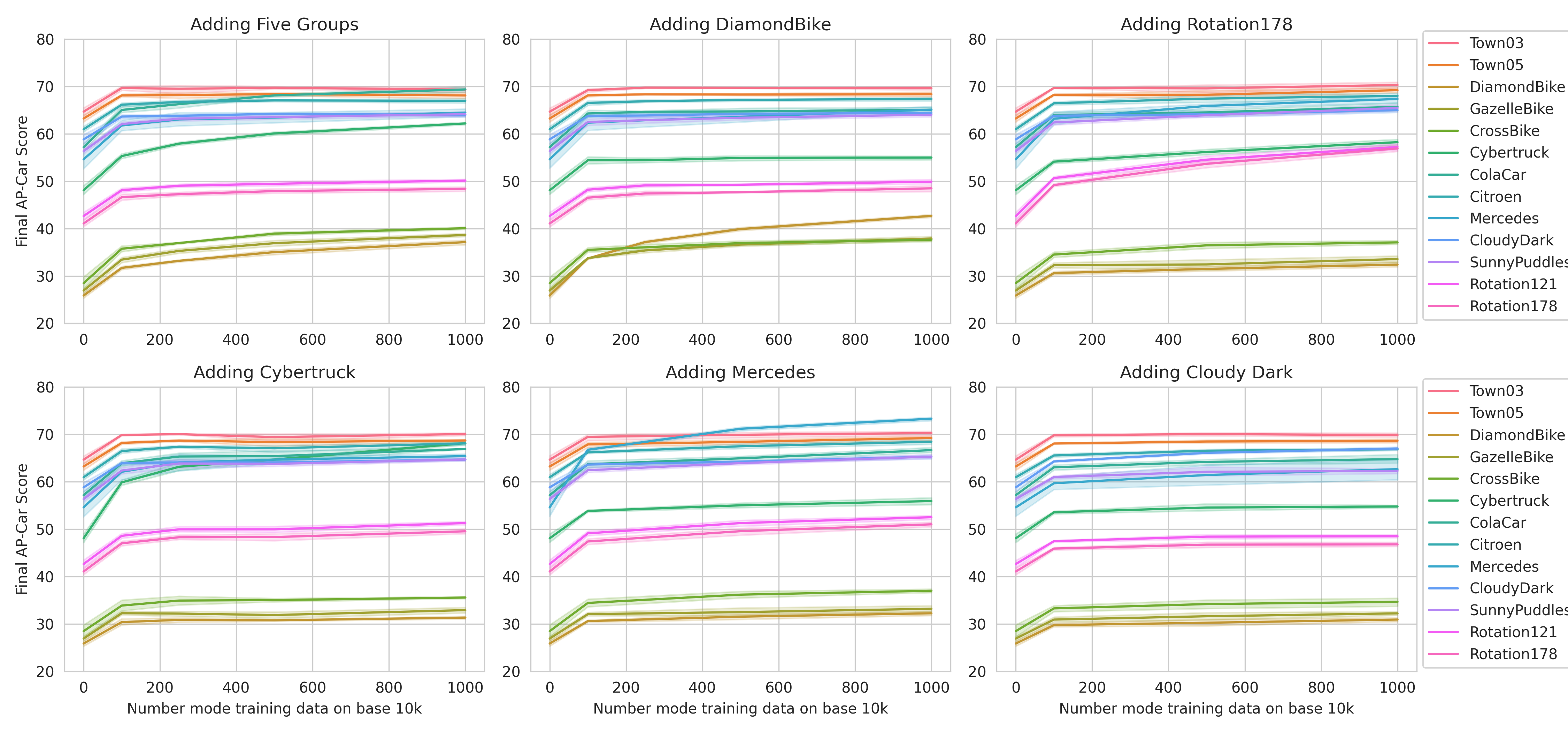}
    \caption{Results adding mode data to the base IID $10000$ training set. This is the same as Figure~\ref{fig:exp3-configs-0-20k} but zoomed into just $[0, 1000]$. The five modes in the top left are the Cybertruck, Cola Car, Diamondback, Gazelle, and Crossbike, each added in equal proportion.}
    \label{fig:exp3-configs-0-1k}%
\end{figure}



\begin{figure}[!h]
    \centering
    \includegraphics[width=.7\textwidth]{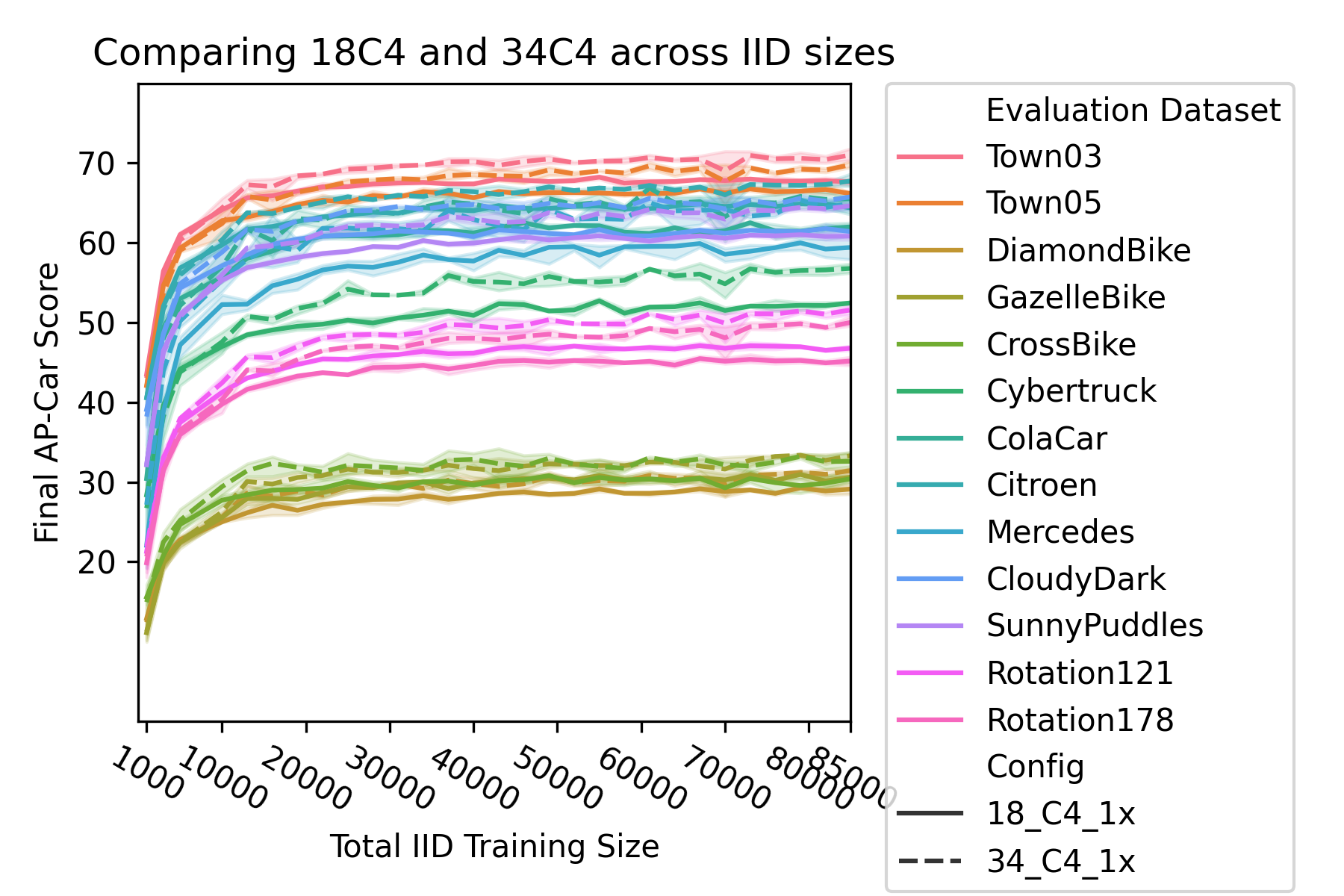}
    \caption{We can see that the model size does matter in that for every group the 34C4 model improves over the 18C4 model. However, the increase is quite small and the data quality and quantity appear to matter much more.}
    \label{fig:bothoverlay}%
\end{figure}

\begin{figure}[!h]
    \centering
    \includegraphics[width=\textwidth]{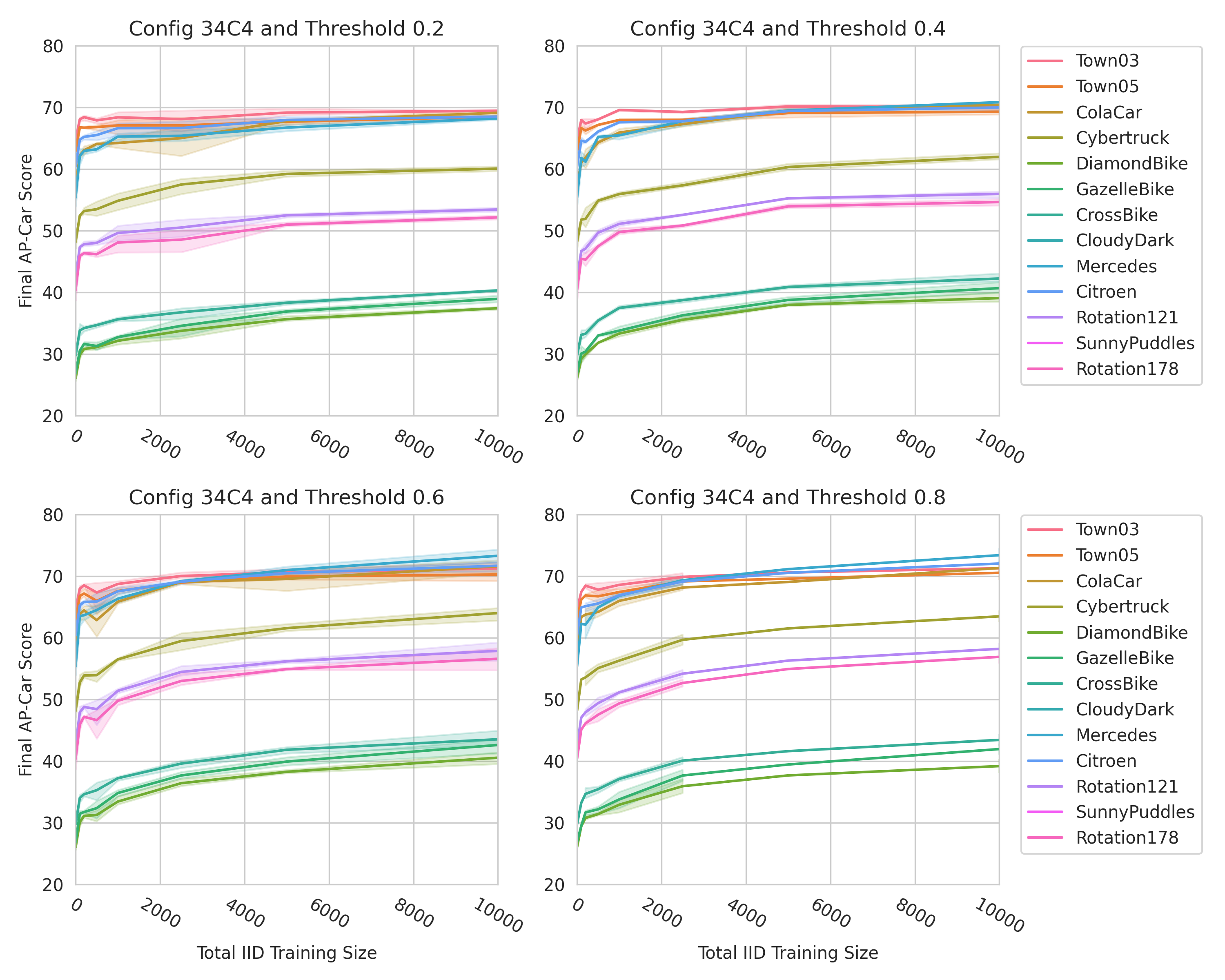}
    \caption{Baseline results training $34\textrm{C}4$ on the base IID $10000$ training set plus additional cause-agnostic data. As specified in Figure~\ref{fig:newexp1-18C}, the additional data is chosen by first selecting a threshold from $[0.2, 0.4, 0.6, 0.8]$, then randomly selecting simulated data for which the model gets at most that score using our scoring function from Section~\ref{sec:method}. This graphic is included for completeness - the results align with what we expect in that they are a little bit better than when using Config $18\textrm{C}4$ for the same task and that they are worse than when performing our proposed method.}
    \label{fig:newexp1-34C}
\end{figure}




\begin{figure}[!h]
    \centering
    \begin{minipage}{0.45\textwidth}
        \centering
        \captionsetup{labelformat=empty}        
        \caption{Ground truth boxes}
        \includegraphics[width=\textwidth]{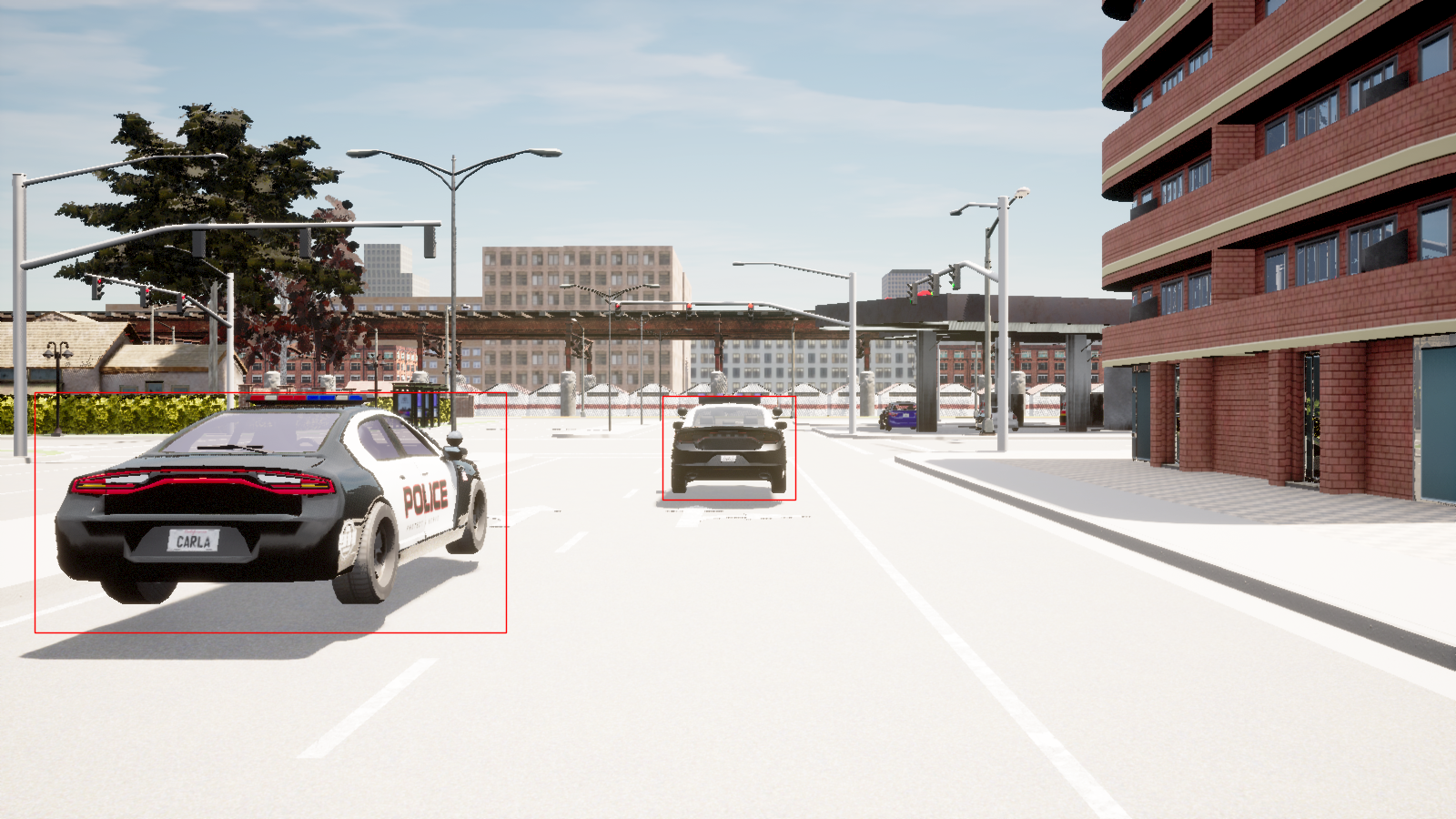} 
    \end{minipage}\hfill
    \begin{minipage}{0.45\textwidth}
        \centering
        \includegraphics[width=\textwidth]{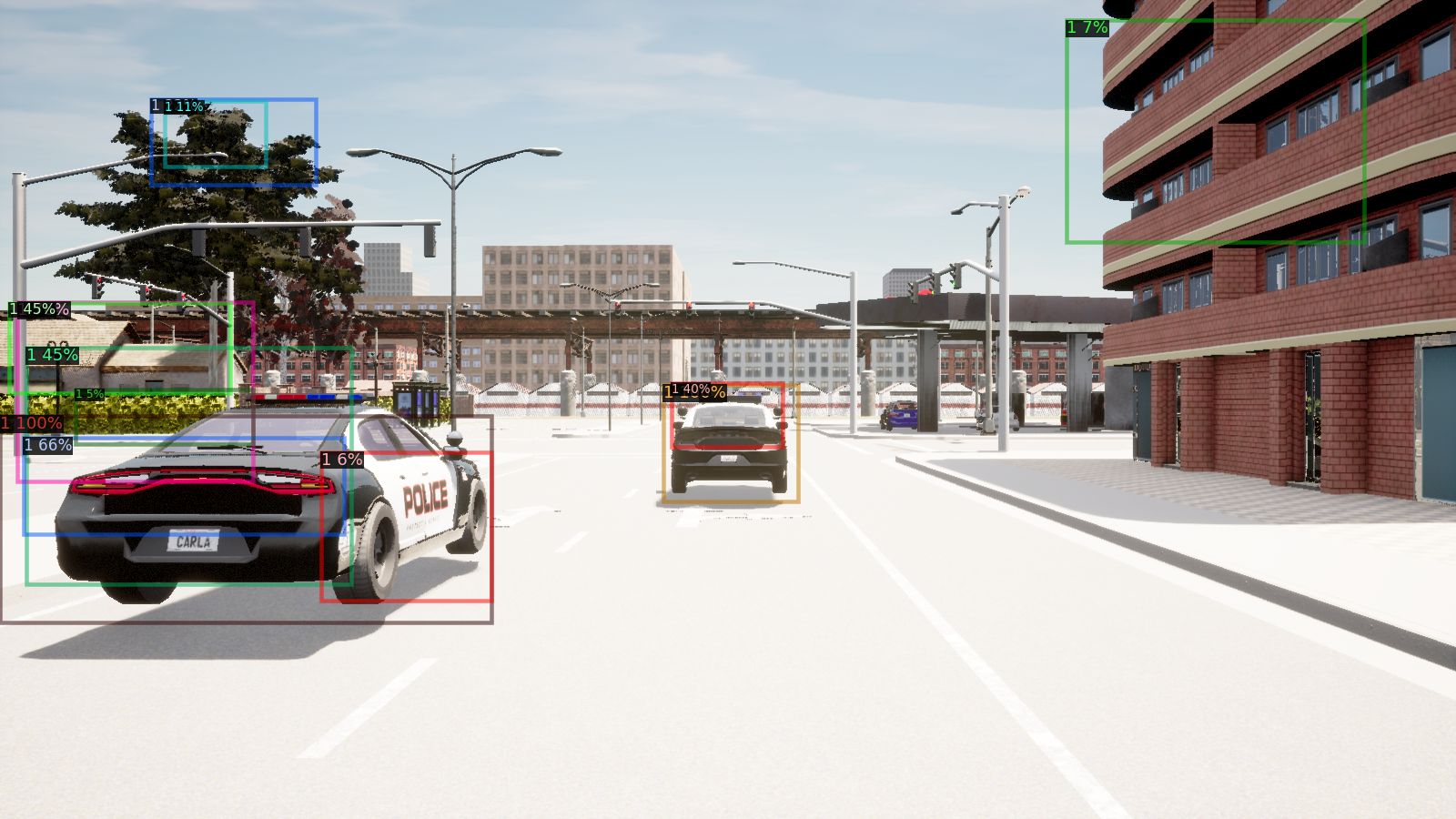} 
    \end{minipage}
    
    \begin{minipage}{0.45\textwidth}
        \centering
        \captionsetup{labelformat=empty}        
        \caption{Object detector's output}
        \includegraphics[width=\textwidth]{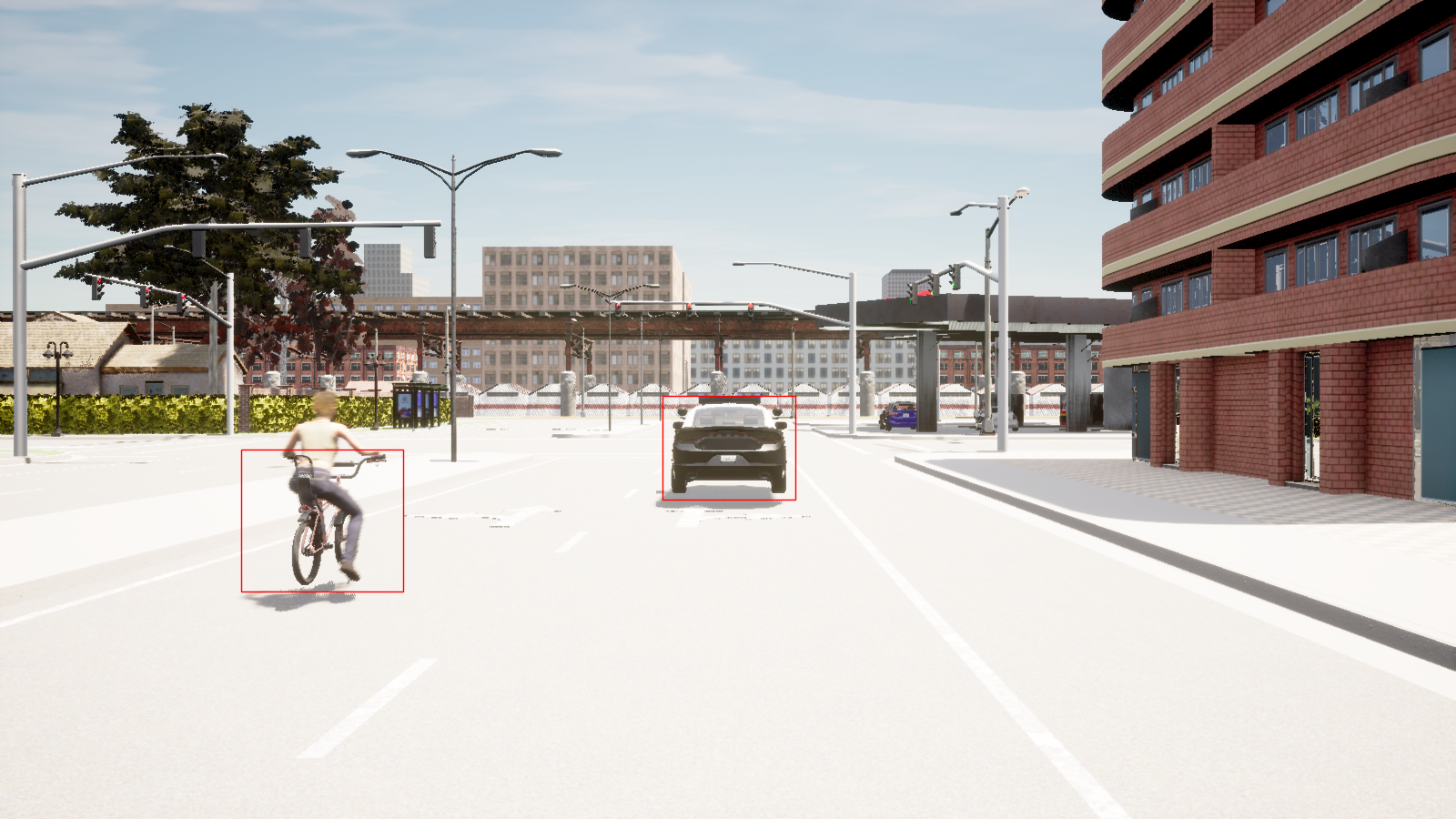} 
    \end{minipage}\hfill
    \begin{minipage}{0.45\textwidth}
        \centering
        \includegraphics[width=\textwidth]{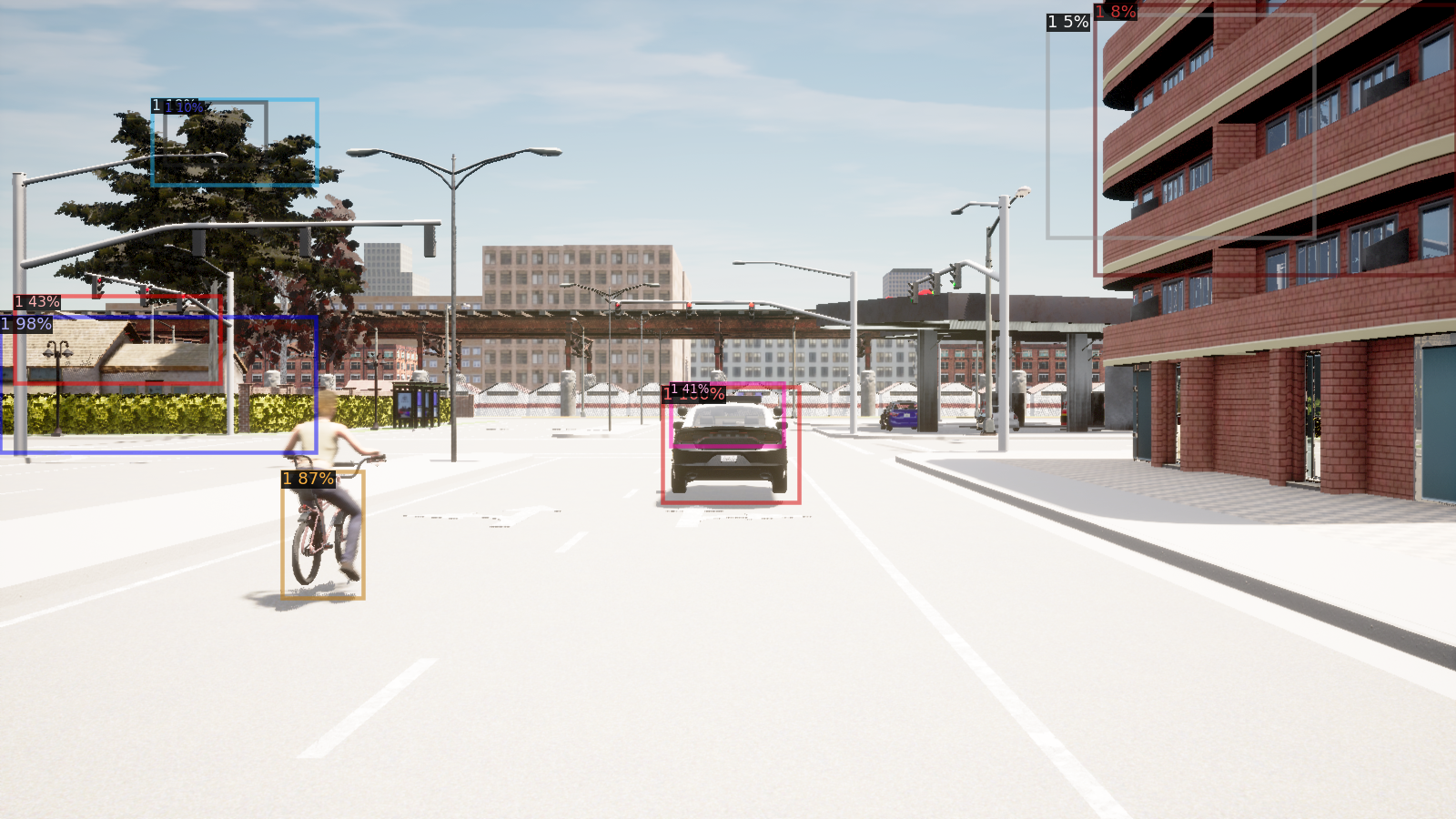} 
    \end{minipage}   
    
    \begin{minipage}{0.45\textwidth}
        \centering
        \includegraphics[width=\textwidth]{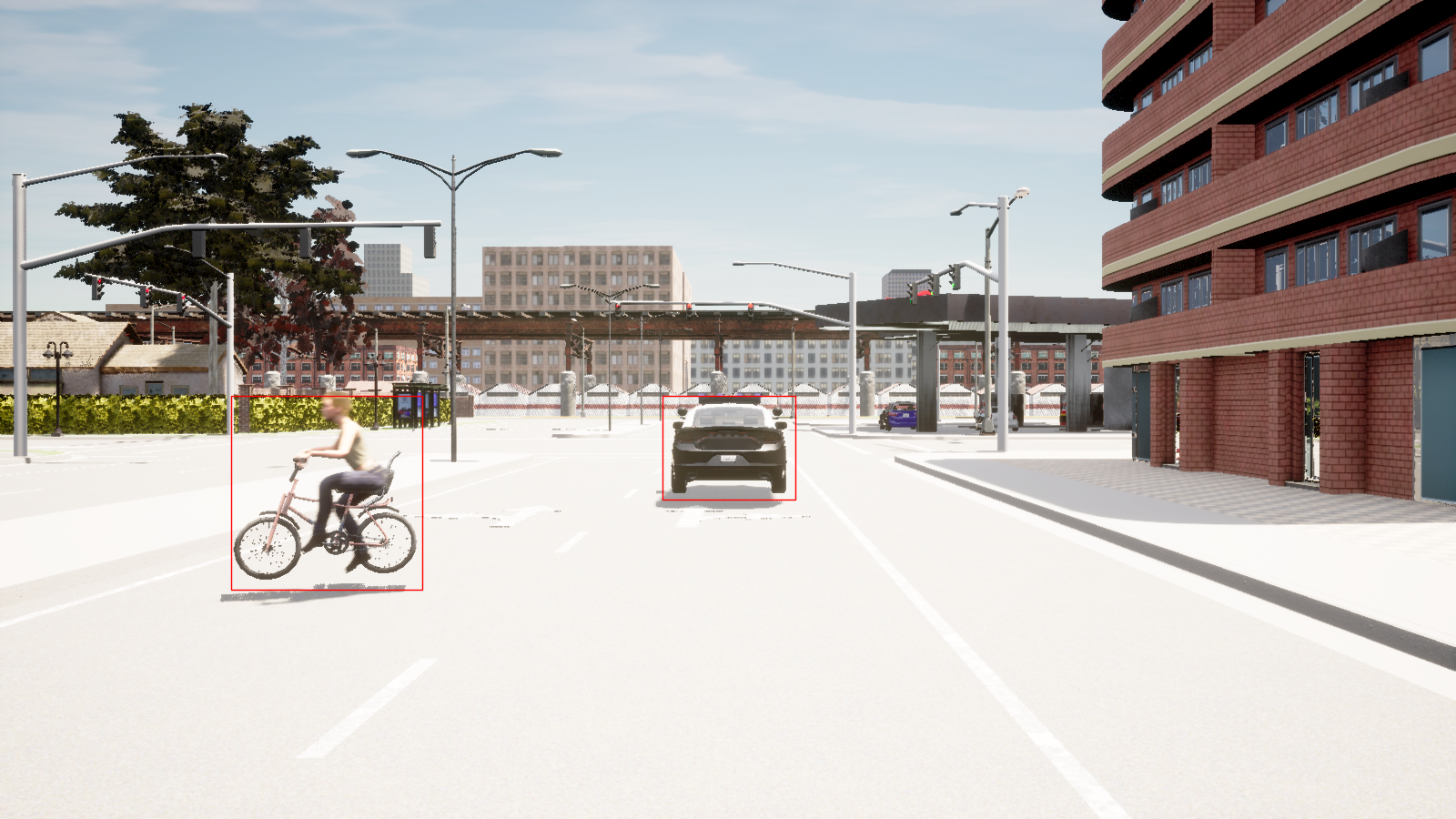} 
    \end{minipage}\hfill
    \begin{minipage}{0.45\textwidth}
        \centering
        \includegraphics[width=\textwidth]{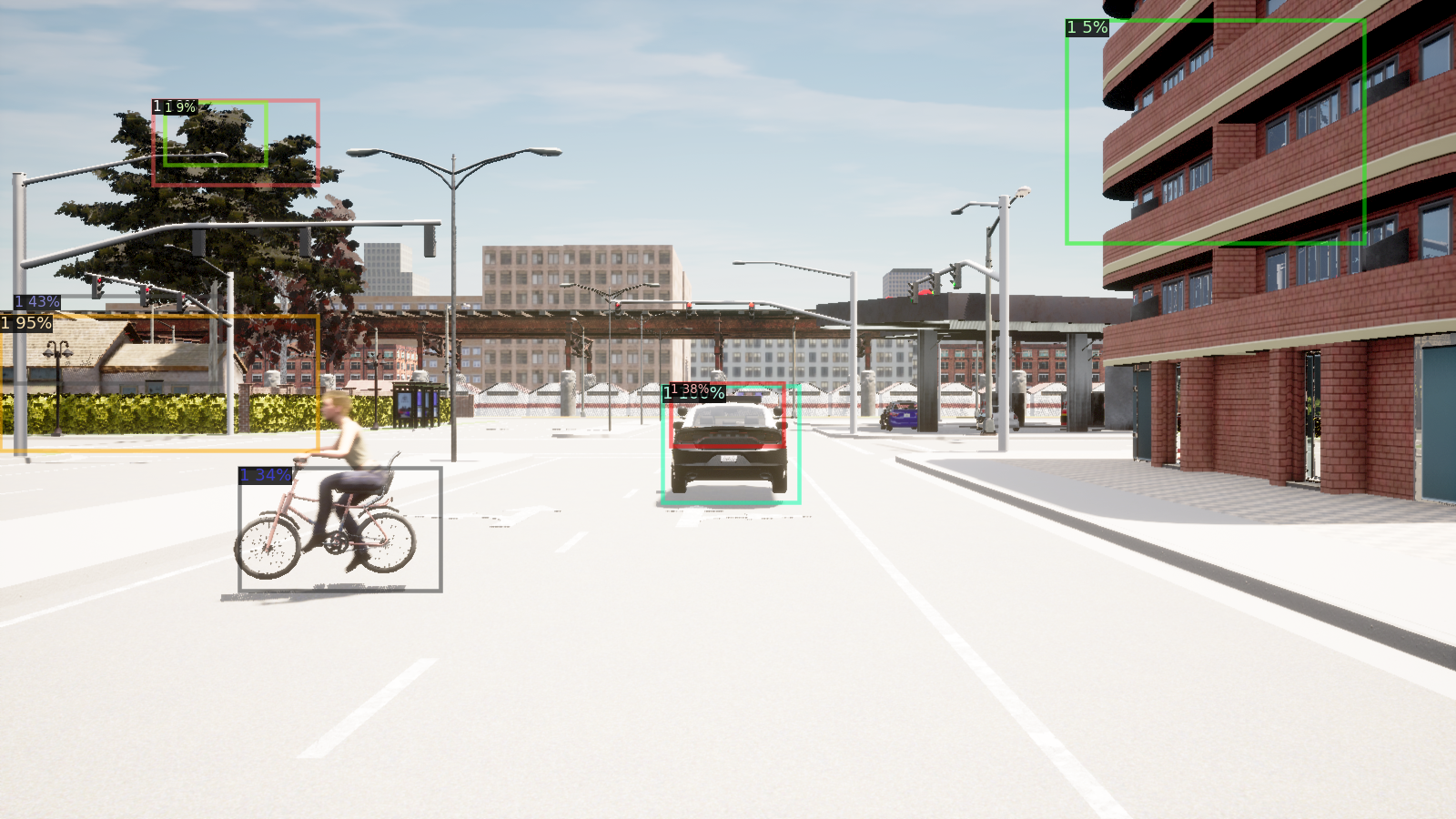} 
    \end{minipage} 
    \caption{\textbf{Interventions taken by the MLM}. The first row is the original scene, the second after an intervention changing the police car to a biker, and the third after an intervention rotating the biker. The left side shows ground truth and the right shows model predictions. The model's predictions were very good for the first scene; in the second scene, it preferred a blank space on the left side to the biker, although the biker did get an $87\%$ confidence. After rotating the biker, that confidence reduces to $34\%$ while the model still hallucinates a vehicle on the left side with $95\%$ confidence.}
    \label{fig:intervention-example-appendix}
\end{figure}

\end{document}